\documentclass{article}

\usepackage{arxiv}

\usepackage[utf8]{inputenc} 
\usepackage[T1]{fontenc}    
\usepackage{hyperref}       
\usepackage{url}            
\usepackage{booktabs}       
\usepackage{amsfonts}       
\usepackage{nicefrac}       
\usepackage{microtype}      

\usepackage{graphicx}
\usepackage{doi}
\usepackage{amsmath}
\usepackage{epsfig}
\usepackage{color}
\usepackage{balance}
\usepackage{hhline}
\usepackage{placeins}
\usepackage{wasysym}
\usepackage{pifont}
\usepackage[nomessages]{fp}
\usepackage{amssymb}
\usepackage{xfrac}
\usepackage{algorithm, algpseudocode, setspace}
\usepackage{multirow}

\newcommand{\norm}[1]{\left\lVert#1\right\rVert}

\newcommand{\Md}{M_\downarrow}

\newcommand{\tM}{\widetilde{M}}

\newcommand{\hMd}{\hat{M}_\downarrow}
\newcommand{\hMda}{\hat{M}_{\downarrow a}}

\newcommand{\hM}{\hat{M}}
\newcommand{\Pd}{P_\downarrow}

\newcommand{\bz}{\mathbf{z}}

\newcommand{\hz}{\mathbf{\hat{z}}}

\newcommand{\ru}{\rule{0mm}{3mm}}

\newcommand{\DL}{D_\lambda}
\newcommand{\DLK}{D_{\lambda}^{(\rm K)}}
\newcommand{\DLV}{\widetilde{D}_{\lambda}^{(\rm K)}}
\newcommand{\DLa}{D_{\lambda, \rm align}^{(\rm K)}}
\newcommand{\DS}{D_{\rm S}}
\newcommand{\DR}{D_{\rho}}

\title{Full-resolution quality assessment for pansharpening}

\author{\href{https://orcid.org/0000-0001-6458-9107}{Giuseppe Scarpa \includegraphics[scale=0.06]{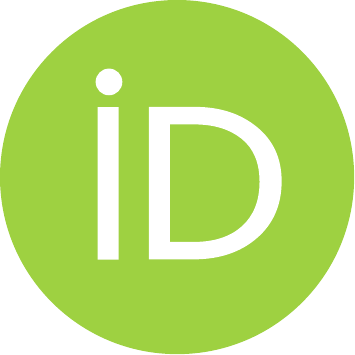}\hspace{1mm}}, \href{https://orcid.org/0000-0001-6577-6879}{Matteo Ciotola \includegraphics[scale=0.06]{orcid.pdf}\hspace{1mm}}\\
	Department of Electrical Engineering and Information Technology (DIETI)\\
	University Federico II\\
	80125 Naples, Italy \\
	\texttt{giscarpa@unina.it},  \texttt{matteo.ciotola@unina.it}\\
}

\date{}

\renewcommand{\headeright}{}
\renewcommand{\undertitle}{}
\renewcommand{\shorttitle}{Full-resolution quality assessment for pansharpening}

\hypersetup{
pdftitle={Full-resolution quality assessment for pansharpening},
pdfsubject={Pansharpening, Evaluation-Metrics},
pdfauthor={Giuseppe Scarpa, Matteo Ciotola},
pdfkeywords={Pansharpening, super-resolution, image enhancement, convolutional neural networks, quality assessment},
}

\begin{document}
\maketitle

\begin{abstract}
	A reliable quality assessment procedure for pansharpening methods is of
critical importance for the development of the related solutions.
Unfortunately, the lack of ground-truths to be used as guidance for an objective evaluation
has pushed the community to resort to two approaches which can also be jointly applied.
Hence, 
two kinds of indexes can be found in the literature:
i) reference-based reduced-resolution indexes aimed to assess the synthesis ability;
ii) no-reference subjective quality indexes for full-resolution datasets aimed to assess spectral and spatial consistency.
Both reference-based and no-reference indexes 
present critical shortcomings which motivate the community to explore new solutions.
In this work, we propose an alternative no-reference full-resolution assessment framework.
On one side we introduce a protocol, namely the reprojection protocol,
to take care of the spectral consistency issue.
On the other side, a new index of the spatial consistency between the pansharpened image and the panchromatic band at full resolution is also proposed.
Experimental results carried out on different datasets/sensors demonstrate the effectiveness of the proposed approach.
\end{abstract}


\section{Introduction}
\label{sec:intro}

Image pansharpening is the process of merging two observations of the same scene, a low resolution multispectral (MS) component and a high resolution panchromatic (PAN) component, to generate a new multispectral image that displays both the rich spectral content of the MS and the high resolution of the PAN.
By following the taxonomy proposed in \cite{Vivone2020}, pansharpening methods can be roughly grouped in four main categories, component substitution (CS) \cite{Shettigara1992},
multiresolution analysis (MRA) \cite{Ranchin2000}, variational optimization (VO) \cite{Vivone2015a, Palsson2020}, and machine/deep learning (ML) \cite{Masi2016, Yang2017}.

In the CS approach,
the multispectral image is transformed in a suitable domain where one of its components is replaced with the PAN.
In the particular case of three spectral bands, 
the Intensity-Hue-Saturation (IHS) transform is an option
where the intensity component can be replaced with the PAN band \cite{Tu2001}.
This method has been generalized in \cite{Tu2004} (GIHS) to handle a larger number of bands.
Other useful transforms to implement a CS solution
include the principal component analysis \cite{Chavez1989},
the Brovey transform \cite{Gillespie1987} and the Gram-Schmidt (GS) decomposition \cite{Laben2000}.
More recently, adaptive CS methods have also been proposed, 
such as the advanced versions of GIHS and GS \cite{Aiazzi2007},
the partial replacement CS method (PRACS) \cite{Choi2011}, or
the band-dependent spatial detail (BDSD) injection method and its variants \cite{Garzelli2008,Garzelli2015, Vivone2019}.

In MRA approaches \cite{Ranchin2000},
the pansharpening task is addressed from the perspective of a pyramidal decomposition to separate low-frequency content from detail components.
The high frequency spatial details are extracted by means of a multi-resolution decomposition, such as
decimated or undecimated wavelet transforms \cite{Nunez1999, Ranchin2000, Otazu2005, Khan2008},
Laplacian pyramids \cite{Aiazzi2002, Aiazzi2003, Aiazzi2006, Lee2010, Restaino2017}, or
other nonseparable transforms, {\it e.g.}, contourlet \cite{Shah2008}.
Extracted details are then properly injected into the upscaled MS component.

A further set of methods address the pansharpening problem through the variational optimization of suitable models of the fused image.
In \cite{Vivone2015a} the optimization target involves the degradation filters mapping high-resolution to low-resolution images,
while \cite{Vicinanza2015} leverage on sparse representations for detail injection. 
Palsson {\it et al.} proposed several methods of this class.
A total variation regularized least square formulation is provided in \cite{Palsson2014}.
The same research team has framed the pansharpening as a maximum {\em a posteriori} problem in \cite{Palsson2015}
and, more recently, explored the use of low-rank representations of the joint PAN-MS \cite{Palsson2020}.
Other methods do not fit the above categories and can be roughly classified as
statistical~\cite{Fasbender2008, Zhang2012, Meng2015, Shen2016, Zhong2017},
dictionary-based \cite{Li2011, Li2013, Zhu2013, Cheng2014, Zhu2016, Hong2019},
or matrix factorization approaches \cite{Yokoya2012, Lanaras2015, Hong2019b}.
The reader is referred to \cite{Vivone2015, Vivone2020} for a more comprehensive review.

In recent years, a paradigm shift from model-based to data-driven approaches has revolutionized all fields of image processing,
from computer vision \cite{Krizhevsky2012, Dong2016, He2017, Lateef2019, Zhao2019} to remote sensing \cite{Yang2017, Scarpa2018, Benedetti2018, Mazza2019}.
In pansharpening,
the first method based on convolutional neural networks (CNN) was proposed by Masi {\it et al.} in 2016 \cite{Masi2016},
after which many more followed in a few years' span \cite{Yang2017, Wei2017a, Wei2017L, Rao2017, Masi2017, Azarang2017, Yuan2018, Liu2018, Shao2018, Vitale2018, Zhan2019,Dong2021A,Dong2021B}.
It seems safe to say that deep learning is currently the most popular approach for pansharpening.
Nonetheless, it suffers from a major problem: the lack of ground truth data for supervised training.
In fact, multi-resolution sensors can only provide the original PAN-MS data, downgraded in space or spectrum,
never their high-resolution versions, which remain to be estimated.

Based on this brief and certainly not exhaustive overview, 
it appears that pansharpening is a very active research field, 
with many new methods proposed every year.
Reliable quality assessment procedures are of critical importance for correctly advancing the state of the art, and a wrong evaluation paradigm may negatively impact the design or tuning of any new solution.
Unfortunately, by the very nature of pansharpening, no Ground Truth (GT) data are available to perform a reference-based assessment. 
As a consequence two kinds of quality assessments are usually employed: 
\begin{itemize}
    \item[{i)}] reference-based reduced-resolution assessment (synthesis check) 
    \item[{ii)}] no-reference full-resolution assessment (consistency check).
\end{itemize}

Lacking GTs, the synthesis capabilities of any pansharpening method can only be assessed on ``synthetic'' data. 
In particular, Wald's protocol \cite{Wald97} suggests taking the real PAN-MS data and applying a proper resolution downgrade process for scale reduction.
The scaled PAN-MS pair is the synthesized image whose GT is now given by the original MS component.
The way the resolution downgrade has to be performed has been object of intense research in the last two decades 
and has a non negligible impact on the reliability of the consequent quality assessment.
However, no matter how a GT is obtained, 
there exist plenty of reference-based image quality indicators.
The spectral angle mapper was introduced in \cite{Kruse1993} and assesses the balance among the spectral bands.
On the contrary, the spatial correlation coefficient \cite{Zhou1998} computes the correlation coefficient across 
the high-pass filtered bands of the pansharpened image and of the GT. 
It is therefore oriented to the assessment of the spatial quality.
The {\em Erreur Relative Globale Adimensionnelle de Synth{\'e}se} \cite{Wald2000} generalizes the root mean squared error, 
introducing band-wise correction weights.
The universal image quality index \cite{Wang2002} compares image statistics to take into account local correlation, intensity and contrast.
Based on this general purpose image quality index, 
domain specific variants suitably adapted to the pansharpening have been proposed in \cite{Alparone2004, Garzelli2009}.
In addition to these indexes, 
other popular general purpose options are sometimes considered for pansharpening,
{\em e.g.}, the peak signal-to-noise ratio or the structural similarity (SSIM) index.

Besides,
spectral or spatial consistency indexes suited for full-resolution images which do not require any GT
have also been developed.
A first spectral distortion index proposed by Zhou {\em et al.} \cite{Zhou1998}
compares a rescaled version of the pansharpened image with the MS image.
Other spectral distortion indexes follow a similar procedure \cite{Alparone2008, Khan2009}.
In \cite{Palubinskas2014} the Quality for Low Resolution (QLR) index based on SSIM was proposed
and later updated exchanging SSIM with the composite image quality measure CMSC based on means, standard deviations and correlation coefficient \cite{Palubinskas2015}.
Basically, most of the known spectral distortion indexes follow this protocol: degradation of the pansharpened image and comparison with the MS image.
The different behavior stems from the degradation model and the error measure. 
For example \cite{Palubinskas2014} is based on the structural similarity indexes while \cite{Palubinskas2015} leverages
on a composite image quality measure based on statistics such as mean, variance and correlation coefficient.
On the other hand, 
the spatial consistency can be checked through the cross-scale invariance of some statistics \cite{Alparone2008}.
A similar approach is proposed in \cite{Khan2009} where only high-frequency components are concerned.
In \cite{Palubinskas2014} it is proposed a Quality for High Resolution (QHR) index which assesses the SSIM index between the panchromatic band and a projection of the pansharpened image in the PAN domain.
A variant of this approach with a different error term was proposed in \cite{Palubinskas2015}.
Another similar approach is proposed in \cite{Alparone2018}, where the coefficient of determination is used to compare the PAN image with its projection from the pansharpened image.
It is also worth to mention other no-reference quality indexes, 
such as the Natural Image Quality Evaluator (NIQE) \cite{Mittal2012, Kwan2017} that seeks to assess the image quality by-itself rather than checking consistency. 

Finally,
often the spectral and the spatial consistency indexes are somehow combined, {\em e.g.} through geometric means, 
to provide a single hybrid consistency index representing the overall quality no-reference assessment of the fused images
\cite{Alparone2008, Khan2009, Palubinskas2014, Palubinskas2015, Aiazzi2014, Kwan2017, Meng2022, Carla2015, Vivone2018c, Vivone2018d}.

Both reference-based and no-reference indexes present inherent limitations which are analyzed in the next Section.

In this work, we propose new full-resolution quality indexes that overcome some of these problems and provide a more reliable guidance for the development of ever more accurate pansharpening methods. To this purpose, 
the developed indexes have been made available to the community through a web repository 
at \url{https://github.com/matciotola/fr-pansh-eval-tool/}.

The remainder of the paper is organized as follows.
Section~\ref{sec:bench} provides a brief critical survey on pansharpening assessment.
Section~\ref{sec:prop} introduces the proposed approach.
Section~\ref{sec:exp} discusses the experimental results and, 
finally, 
Section~\ref{sec:conclusions} draws conclusions.

\section{A review of panshapening indexes}
\label{sec:bench}

Due to the lack of ground truths, visual inspection by human experts is the ultimate benchmark for pansharpening methods.
However, this is a lengthy and tedious work, and numerical indexes are essential for a viable development process.
Reduced-Resolution (RR) indexes are computed, according to Wald's protocol \cite{Wald97}, by reducing the scale of both MS and PAN components. Then, pansharpening is performed on these rescaled data, and the original MS is used as GT to compute reference-based indexes.
Instead, Full-Resolution (FR) indexes are computed on the original data and try to assess their spectral and/or spatial consistency with the pansharpened output.
In a comprehensive evaluation, following Wald's protocol \cite{Wald97}, 
usually, both types of indexes are considered together in addition to visual inspection \cite{Vivone2020}.
In order to highlight the critical points of the current approaches to the quality assessment, 
we provide a brief overview of some of the most representative indexes in the following.

\subsection{Reduced-resolution assessment}
Let $M$ and $P$ be the original MS and PAN components, respectively, and $\hM$ be the pansharpened image.
The synthesis properties can not be directly verified on the full-resolution pair $(M,P)$ due to the lack of the GT with which to compare $\hM$. 
Leveraging on a scale invariance hypothesis, 
it is however possible to synthesize datasets with GT by properly downgrading the available $(M,P)$ pairs by a factor $R$ (PAN-MS resolution ratio). 
The resulting reduced resolution pair $(\Md,\Pd)$ can therefore be used as the source input for the pansharpening task,
whereas the original MS image $M$ plays as GT.
Once moved to the reduced resolution space, 
many different quality indexes can be computed to assess the mismatch between the pansharpened image 
and its reference GT, such as root mean square error, structural similarity, and so on.
In particular, referring to some of the most popular ones for pansharpening, the most common are the following.

\begin{itemize}
    \item[{a)}] SAM (Spectral Angle Mapper) \cite{Kruse1993}. It determines the spectral similarity in terms of pixel-wise 
    average angle between spectral signatures. Said $\mathbf{v}$ and $\mathbf{\hat{v}}$ two corresponding 
    pixel spectral responses to be compared, SAM is obtained by averaging over all image locations the following
    ``angle'' among vectors:

   \begin{equation}
   		{\rm SAM}(\mathbf{v},\mathbf{\hat{v}}) = 
   				\arccos\left( \frac{\left\langle \mathbf{v},\mathbf{\hat{v}} \right\rangle}{\|\mathbf{v} \|_2 \cdot \| \mathbf{\hat{v}} \|_2}\right).
   \end{equation}

    \item[{b)}] ERGAS ({\em Erreur Relative Globale Adimensionnelle de Synth{\'e}se})  \cite{Wald2000}. 
    It is one of the most popular indexes to assess both spectral and structural fidelity 
    between a synthesized image and a target GT.
	   Such an index presents interesting invariance properties. 
	   Indeed, it is insensitive to radiometric range, number of bands and resolution ratio.
	   Said $B$ the number of spectral bands, it is defined as

	   \begin{equation}
	   {\rm ERGAS}  = \frac{100}{R} \sqrt{\frac{1}{B} \sum_{b=1}^B \left(\frac{{\rm RMSE}_b}{\mu^{\rm GT}_b}\right)^2},
	   \end{equation}
	   where ${\rm RMSE}_b$ is the root mean square error over the $b$-th spectral band
	   and $\mu^{\rm GT}_b$ is the average intensity of the $b$-th band of the GT image.

	\item[{c)}] $Q2^n$ \cite{Garzelli2009}. It is a multiband extension of the universal image quality index (UIQI) \cite{Wang2002}.
	Each pixel of an image with $B$ spectral bands is accommodated into a hypercomplex (HC) number 
	with one real part and $B$– 1 imaginary parts. Let $\bz$ and $\hz$ denote the HC representations of a generic 
	GT pixel and its prediction, respectively, 
	then $Q2^n$ can be written as the product of three terms:

	\begin{equation}
	Q2^n = \frac{|\sigma_{\bz\hz}|}{\sigma_\bz \sigma_\hz} 
				\cdot \frac{2 \sigma_\bz \sigma_\hz}{\sigma_\bz^2 + \sigma_\hz^2} 
				\cdot \frac{2 \mu_\bz \mu_\hz}{|\mu_\bz|^2+|\mu_\hz|^2}. 
	\end{equation}
	The first factor provides the modulus of the HC correlation coefficient between $\bz$ and $\hz$. 
	The second and the third terms measure contrast changes and mean bias, respectively, on all bands simultaneously. 
	Statistics are typically computed on 32$\times$32 pixels blocks, then and an average over the blocks of the whole image
	provides the global $Q2^n$ score, which takes values in the $[0,1]$ interval, being 1 the optimal value achieved if and 
	only if $\bz=\hz$ in each location.	
\end{itemize}

Behind its appealing simplicity, 
regardless of the specific error measurement employed,  the reduced resolution assessment 
hides some major subtle pitfalls that undermine its usefulness.
On one hand, its accuracy depends critically on the method used for scaling, 
which may not correspond to the actual sensing conditions.
More fundamentally, it relies on the arbitrary assumption that a method optimized for the RR scale will keep its good behavior at the FR scale.
Experimental evidence shows this not to be the case.
Furthermore, 
optimizing parameters for a given scale may lead to a sort of ``scale overfitting'', with the perverse effect of degrading performance on a different scale.
For these reasons, many recent studies focus on no-reference full-resolution quality indexes \cite{Kwan2017, Alparone2018, Vivone2018c, Meng2022}.
It is also worth to recall that the assessment of spectral and spatial consistencies is not necessarily linked to a fusion task
and is of a broader interest, with applications in the medical \cite{Szczykutowicz2021}, agricultural \cite{Kordi2021} and food \cite{Zhu2020} sciences, to mention a few.

\subsection{Full-resolution no-reference assessment}
As a consequence of the above limitations, 
many studies have moved toward a {\em no-reference} approach to the pansharpening quality assessment.
In particular, the same Wald's protocol \cite{Wald97} recognizes the problem invoking a complementary {\em consistency} quality check in addition to the synthesis capacity assessment.
In practice, 
for consistency-based assessment, 
the input data are pansharpened in their original (full) resolution space. 
The outcome is then compared with both input components, MS and PAN, 
for spectral and consistency assessment, respectively.
The spectral consistency check requires a spatial resolution degradation of the pansharpened image, 
which is typically done by means of a Low-Pass Filtering (LPF) followed 
by decimation, using LPFs that approximate the sensor Modulation Transfer Function (MTF) \cite{Alparone2007}.
Other approaches have also been explored,
such as wavelet transforms \cite{Daubechies1993, Ranchin1993}, averaging operators \cite{Munechika1993}, 
pyramid decomposition \cite{Aiazzi2002}.
The spatial (or {\em structural}) consistency check is indeed more controversial
lacking a clear relationship that allows to project the pansharpened MS in the PAN domain \cite{Thomas2008}. 
No-reference indexes are also often referred to as FR indexes, 
as they do not require any resolution degradation of the input data, 
contrarily to the reference-based ones which do need it because of the lack of GTs.
For this reason, the latter are also referred to as RR indexes.

Among the several FR quality assessment approaches proposed in the last years \cite{Alparone2008,Zhou1998, Khan2009,Shah2008,Alparone2018},
we will recall a few of them which are the most frequently used.
In particular, Alparone {\em et al.} \cite{Alparone2008} proposed a spectral distortion index $\DL$ defined as

\begin{equation}
\DL = \sqrt[p]{\frac{1}{B(B-1)}\sum_{i=1}^B \sum_{j=1, j\neq j}^B
					\left| d_{i,j}(M,\hM) \right|^p},
\end{equation}
where $d_{i,j}(M,\hM) = Q(M_i,M_j) - Q(\hM_i,\hM_j)$,
with subscripts indicating selected bands and being $Q(\cdot,\cdot)$  the UIQI \cite{Wang2002}.
Such an index was also enclosed in the pansharpening benchmark toolbox \cite{Vivone2015}.
As it can be observed,
$\DL$ compares the inter-band relationship through the UIQI, 
separately for the input MS $M$ and the pansharpened image $\hM$,
and then quantifies their divergence over each coupling of spectral bands.
As it can be figured, 
lacking a direct measurement of the distance between $M$ and $\hM$,
and leveraging on an assumption of cross-scale inter-band correlation invariance only, 
there can be important spectral aberrations that can remain unseen by $\DL$.
The recently published revisited version \cite{Vivone2020} of the same toolbox \cite{Vivone2015}, indeed, 
makes use of an approximated version of Khan's index proposed in \cite{Khan2009} to monitor the spectral consistency.
In particular, 
Khan's index is defined as follows

\begin{equation}
\DLK = 1-Q2^n \left(\hMd^{\rm lp}, M  \right),
\label{eq:dlk}
\end{equation}
where $\hMd^{\rm lp}$ indicates the MTF-based low-pass filtered and decimated version of the pansharpened image $\hM$, 
while $M$ is the original input MS.
Its approximation used in the toolbox \cite{Vivone2020},
name it $\DLV$,
avoids the decimation of $\hM$ by using an upscaled version $\tM$ of $M$:

\begin{equation}
\DLV = 1-Q2^n \left(\hM^{\rm lp}, \tM  \right),
\label{eq:dl}
\end{equation}
being $\hM^{\rm lp}$ the low-pass version of $\hM$.
These indexes range between 0 (optimal value) and 1, and clearly relate to the spectral consistency, 
since the resolution downgrade process applied to $\hM$ gets rid of the high-pass content retrieved from the PAN.
However, two critical points should be taken into account:
\begin{itemize}
\item[{a)}] dependence on the accuracy of the estimated MTF;
\item[{b)}] sensitivity to the PAN-MS alignment.
\end{itemize}

In particular, 
as a consequence of b),
methods such as several CS approaches \cite{Lolli2017, Laben2000, Aiazzi2007, Restaino2017, Choi2011} 
that intrinsically address the coregistration problem are penalized by both $\DLK$ and $\DLV$ on misaligned datasets.
Hence, in order to fairly use Khan's index it is usually recommended to keep separate the registration and pansharpening problems
by means of a prior coregistration of the datasets to be used for testing \cite{Vivone2015,Vivone2020}.

Several indexes have also been proposed for FR spatial consistency check \cite{Alparone2008, Khan2009, Palubinskas2014, Alparone2018}.
In particular, the spatial distortion index proposed in \cite{Alparone2008}
is used in the most recent pansharpening assessment toolbox \cite{Vivone2020}
and is defined as:

\begin{equation}
\DS = \norm{\frac{1}{B} \sum_{i=1}^B Q(\hM_i,P) - Q(M_i, \Pd)}_{q},
\label{eq:ds}
\end{equation}
where $\Pd$ is the original PAN component resized to the MS scale and $Q(\cdot,\cdot)$ is the UIQI quality index \cite{Wang2002}.
This index is somehow similar to $\DL$ 
as it checks the cross-scale invariance of the PAN-MS spectral correlation.
Therefore it inherits similar limitations, which are:
\begin{itemize}
\item[{i.}] no direct comparison between the pansharpened image $\hM$ and the PAN $P$;
\item[{ii.}] a cross-scale invariance assumption for which there are no guarantees.
\end{itemize}


\section{Proposed full-resolution indexes}
\label{sec:prop}

We now propose some new indexes for assessing the quality of pansharpening, aiming to overcome some limitations of traditional ones.
Given the weakness of the scale-invariance assumption of reduced-resolution quality assessment indexes, we focus on full-resolution ones,
moving our perspective from synthesis to consistency.
In the following, we will discuss spectral and spatial assessment separately.

\subsection{Reprojection protocol for spectral accuracy assessment}

\newcommand{\M}{\mathcal{M}}
The assessment of the spectral consistency is not 
as controversial as the assessment of the spatial one.
In particular, as far as we know, 
Khan's approach is the most trusted one for this purpose.
Nonetheless, as remarked above, it has some limitations and, 
in particular,
in order to work properly, 
the datasets must be well aligned. 

To cope with this issue without the need to operate any coregistration on the available datasets,
let us now consider a variation of Khan's index where the decimation of $\hM^{\rm lp}$ 
is performed using a band-wise shift that maximizes the correlation coefficient 
between the low-pass filtered PAN $P^{\rm lp}$ and each ideally interpolated MS band $\tM_b$.
In case of aligned datasets, 
it returns the usual Khan's index because no shifts are applied.
Otherwise, before decimation, we apply to $\hM$ 
the same shift that would align $\tM$ and $P^{\rm lp}$.
Let us indicate as $\DLa$ such an index which is formally given by

\begin{equation}
\DLa = 1-Q2^n \left(\hMda^{\rm lp}, M  \right),
\label{eq:DLa}
\end{equation}
where ${\downarrow}{a}$ indicates decimation with band-wise alignment.

Let $\M$ indicate a generic reference-based distortion index, such as $Q2^n$, SAM or ERGAS.
We define the reprojection index R-$\M$ as

\begin{equation}
\mbox{R-}\M = \M\left(\hM^{\rm lp}_{\downarrow a},M \right),
\label{eq:RM}
\end{equation}
that in the particular case of $\M=Q2^n$ 
reduces to the complement of the $\DLa$ variant defined above (Eq.~\ref{eq:DLa}), {\em i.e.},

\begin{equation} 
	\mbox{R-}Q2^n = Q2^n \left(\hMda^{\rm lp}, M  \right) = 1- \DLa.
\end{equation}
In other words,
the pansharpened image $\hM$ is projected in the original MS space prior to be compared with the MS image $M$
(reference) using the error index $\M$.
The projection accounts for both eventual misalignment and sensor MTF using matched low-pass filters.
Of course,
because of the low-pass filtering, 
the high frequency image content (spatial details) 
is discarded and some further indexes will be necessary to assess spatial quality.
Note also that,
under the hypothesis that the sensor MTF is correctly simulated, 
the reprojection indexes ensure that an ideal pansharpening (GT) would achieve the best score as expected.
This will be experimentally analyzed in Sec.~\ref{sec:cross-check}. 

\begin{figure}
\centering
\includegraphics[width=0.8\columnwidth]{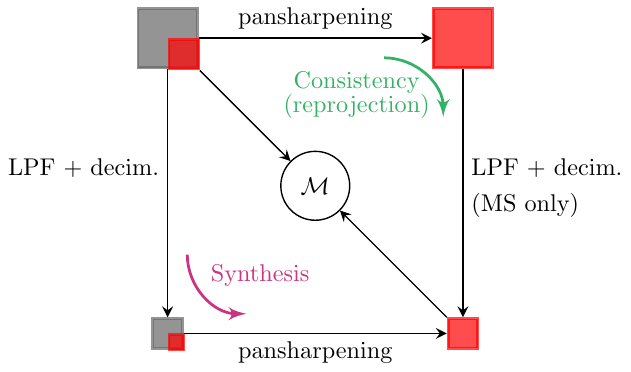}
\caption{Spectral consistency check using reprojection (clockwise) {\em vs} synthesis check according to Wald's protocol (counterclockwise).
$\mathcal{M}$ is any reference-based index. In gray and red are shown the PAN and MS channels, respectively.}
\label{fig:R-SAM}
\end{figure}

A pictorial representation of the proposed reprojection indexes which relates them to the synthesis assessment protocol proposed by Wald \cite{Wald97}
is given in Fig.\ref{fig:R-SAM}.
The common starting point (top-left) is the available $(P,M)$ pair.
Indexes based on Wald's protocol follow the counterclockwise path: decimation-pansharpening-measure.
The proposed reprojection protocol, instead,
consists in following the clockwise path: pansharpening-decimation-measure,
where the actual pansharpened image is first computed, and then reprojected on the low-resolution domain for quality assessment.
Eventually, in both cases, the original MS is used as a reference for any error measurement.
In our proposal, however, pansharpening precedes the resolution downgrading, thereby exploiting the most valuable source of spatial information, the original PAN.
Furthermore, the overall evaluation no longer depends on how the PAN is downscaled.
Since the the reprojection protocol involves only the low-resolution component of the synthesized image,
it has to be considered a spectral consistency rather than a synthesis check.
A pseudo-code of the reprojection indexes is given in Algorithm~\ref{algo:spectmarker}.
\newcommand{\hP}{P^{\rm lp}}
\begin{algorithm}
\footnotesize
\begin{spacing}{1.1}
\begin{algorithmic}[1]
\Require{$\hM$, $M$, $P$, $\mathbf{g}_{\rm MTF}$ (MTF parameters), $\mathcal{M}$ (error function)}
\Ensure{R-$\mathcal{M}$}
\State $\tM = {\rm upscale}(M, R)$ \Comment polynomial interpolation
\State $\hP = {\rm LPF}(P,\mathbf{g}_{\rm MTF})$ \Comment MTF-tailored low-pass filtering
\For{$b=1:B$}
    \State $\mathbf{s}(b) =  \underset{(m,n) \in [1,R]^2}{\arg\max} \mbox{corrcoef}\left( \hP_{i,j}, \tM(b)_{i+m, j+n} \right)$     \Comment band-wise shift for alignment 
\EndFor

\State $\hM^{\rm lp} = {\rm LPF}(\hM, \mathbf{g}_{\rm MTF})$
\State $\hMda^{\rm lp} = {\rm decimation}(\hM^{\rm lp}, R, \mathbf{s})$
\State $\mbox{R-}\M = \M\left(\hM^{\rm lp}_{\downarrow a},M \right)$ \Comment Eq. \ref{eq:RM}
\end{algorithmic}
\end{spacing}
\caption{Reprojection error assessment}
\label{algo:spectmarker}
\end{algorithm}

\subsection{Correlation-based spatial consistency index}
\label{sec:rho}

In order to complement the reprojection-based assessment, 
we also propose a new index to evaluate the spatial consistency
with the aim of obtaining indications better correlated to human judgment than those provided by $\DS$.
In particular, to assess the preservation of fine-scale spatial structures,
we compute the average local correlation between the pansharpened image and the PAN.

Let $X^\sigma_{ij}$ indicate a $\sigma\times\sigma$ patch of $X$ centered on location $(i,j)$.
We compute the correlation field $\rho^{\sigma}_{P\hM}$ given by the local correlation coefficients between $P$ and each band $b$ of $\hM$:

\begin{equation}
    \rho^{\sigma}_{P\hM}(i,j,b) = \mbox{corrcoef}\left( P^\sigma_{ij}, \hM(b)^\sigma_{ij} \right)
\label{eq:localcorrcoef}
\end{equation}
Then we reduce it to its average value $\overline{\rho^{\sigma}_{P\hM}}$ over space and spectral bands.
The final index is then defined as

\begin{equation}
    \DR \triangleq 1 - \overline{\rho^{\sigma}_{P\hM}},
\label{eq:Xdist}
\end{equation}
such that $\DR=0$ corresponds to perfect correlation.
From a different point of view,
we are studying to what extent a $\sigma{\times}\sigma$ patch of any band of $\hM$ can be linearly predicted from the corresponding PAN patch.
Therefore, $\DR$ is strictly related to the matching between the spatial layouts of $\hM$ and $P$ at a fine scale.

The choice of the scale parameter $\sigma$ is of critical importance,
as we are interested to the exploitation of the complementary information discarded by the reprojection indexes, 
{\em i.e.} details appearing only at the PAN scale but not at the MS scale.
Therefore, 
if $R$ is the resolution ratio,
by choosing $\sigma = R$ (equal to 4 for our datasets) we can put the focus just on that spatial content which is not visible at the MS scale.
By doing so, we reduce the dependence between the proposed spatial ($\DR$)  and spectral (R-$\M$) distortion indexes,
assigning to them clear but well separated evaluation objectives.

As for $\DS$ and $\DLK$, or its variants,
$\DR$ computed for $\hM$=GT does not reach the zero value,
given the slightly different spatial layout of different bands in natural images and the complex relationship between the high resolution image and its panchromatic projection \cite{Thomas2008}.
Nonetheless, we expect good quality pansharpened images to present relatively small values of $\DR$.
A detailed discussion in this regard supported by our experimental results will be provided in the following.

\section{Experimental results and discussion}
\label{sec:exp}

In this Section we present several experimental analyses. 
Preliminarily, 
a summary of the datasets employed and on the involved methods is provided.
The first experiment focuses on the dependence of the spectral consistency indexes on the alignments of the MS bands with the PAN.
Next, we move to the reduced-resolution domain to cross-check reference-based and no-reference indexes thanks to the availability of the ground-truth.
Then, it follows an experimental analysis focused on the assessment of the spatial consistency,
before closing with an overall comparison.

\subsection{Datasets and methods}
\begin{table}[b]
\centering
\caption{Bandwidths of the multispectral channels (left-hand side) and Ground Sample Distance (GSD) [m] at Nadir (rightmost column) for WorldView-2 and WorldView-3 images.}
\small
\setlength{\tabcolsep}{2pt}
\newcommand{\vsp}{\rule{0pt}{10pt}}
\begin{tabular}{l@{\rule{7pt}{0pt}}cccccccc@{\rule{7pt}{0pt}}c}
\toprule
\vsp \textbf{Sensor} & \multicolumn{8}{c}{\textbf{Bandwidths of the MS Channels [nm]}} & \textbf{GSD [m]}\\
\vsp      & \textbf{Coastal} & \textbf{Red}        & \textbf{Blue}        & \textbf{Red Edge} & \textbf{Green}   & \textbf{Near-IR1} & \textbf{Yellow} & \textbf{Near-IR2} & \textbf{PAN/MS} \\
            \midrule
\vsp WV2   &  396-458 & 624-694 & 442-515 & 699-749 & 506-586 & 765-901 & 584-632 & 856-1043 & 0.46 / 1.84\\
\vsp WV3   &  400-450 & 630-690 & 450-510 & 705-745 & 510-580 & 770-895 & 585-625 & 860-1040 & 0.31 / 1.24\\
        \bottomrule
\end{tabular}
\label{tab:sensors}
\end{table} 

The experimental validation relies on 25 methods provided in the benchmark toolbox \cite{Vivone2020}
belonging to the four main categories recalled in Section~\ref{sec:intro}, CS (8), MRA (9), VO (3) and ML (4), plus an ideal interpolator (EXP).
The dataset is composed by two WorldView-2 (WV2) and two WorldView-3 (WV3) large images, courtesy sample products of DigitalGlobe$^\copyright$.
20 $512{\times}512$ tiles were extracted from the WV2 images (Washington and Stockholm) and 20 from the WV3 images (Adelaide and Mexico City) for the experiments at full resolution.
Likewise, 20+20 $2048{\times}2048$ tiles were extracted and downscaled to size $512{\times}512$ for the experiments at reduced resolution.
Tab.~\ref{tab:sensors} summarizes the main spectral and spatial characteristic of the WV-2 and WV-3 sensors.

\subsection{Spectral distortion dependence on PAN-MS misalignment}

\begin{table}
\centering
\caption{Misregistration impact on $\DL$, $\DLK$, $\DLV$ and $\DLa$. Detailed information about the methods can be found in \cite{Vivone2020}.}
\footnotesize
\setlength{\tabcolsep}{2pt}
\begin{tabular}{ll@{\rule{7mm}{0mm}}cc@{\rule{7mm}{0mm}}cc@{\rule{7mm}{0mm}}cc@{\rule{7mm}{0mm}}cc}
\hline \hline
&  &\multicolumn{2}{c}{\ru $\DL${\rule{7mm}{0mm}}}     &    \multicolumn{2}{c}{$\DLK${\rule{7mm}{0mm}}} &   \multicolumn{2}{c}{$\DLV${\rule{7mm}{0mm}}} & \multicolumn{2}{c}{$\DLa$} \\
& Aligned dataset  & yes  & no & yes & no & yes & no & yes & no \\
\hline \hline
& \ru EXP	(interpolator)		  & 0.0000 &    0.0000 & 0.0249 &   0.0336 &	0.0394 &	0.0479 & 0.1238 & 	0.0670\\  \hline
\multirow{8}{*}{CS} & \ru BT-H \cite{Lolli2017}		  & 0.0698 &    0.0823 & 0.1231 &   0.0696 & 	0.2310 &	0.1324 & 0.0874 &	0.0697\\
&\ru BDSD \cite{Garzelli2008}		  & 0.0429 &	0.0377 & 0.0989 &	0.0867 &	0.1834 &	0.1561 & 0.1511 &	0.1064\\
&\ru C-BDSD \cite{Garzelli2015}		  & 0.0514 &	0.0557 & 0.1436 &	0.1195 &	0.2158 &	0.1836 & 0.2346 &	0.1664\\	
&\ru BDSD-PC	\cite{Vivone2019} 	  & 0.0141 &	0.0160 & 0.0782 &	0.0896 &	0.1464 &	0.1530 & 0.0705 &	0.0892\\	
&\ru GS \cite{Laben2000}			  & 0.0196 &	0.0177 & 0.1494 &	0.0824 &	0.2611 &	0.1414 & 0.0824 &	0.0839\\
&\ru GSA	 \cite{Aiazzi2007}		  & 0.0576 &	0.0573 & 0.1182 &	0.0604 &	0.2457 &	0.1334 & 0.0760 &	0.0624\\
&\ru C-GSA \cite{Restaino2017}	      & 0.0309 &	0.0333 & 0.0929 &	0.0583 &	0.1925 &	0.1240 & 0.0741 &	0.0625\\
&\ru PRACS	  \cite{Choi2011}      & 0.0178 &	0.0193 & 0.0728 &	0.0468 &	0.1464 &	0.0889 & 0.0834 &	0.0610\\ \hline
\multirow{9}{*}{MRA} &\ru AWLP \cite{Alparone2017}	      & 0.0332 &	0.0416 & 0.0282 &	0.0273 &	0.0415 &	0.0320 & 0.0920 &	0.0495\\
&\ru MTF-GLP	 \cite{Alparone2017}     & 0.0679 &	0.0759 & 0.0231 &	0.0195 &	0.0434 &	0.0352 & 0.0912 &	0.0428\\
&\ru MTF-GLP-FS \cite{Vivone2018a}	  & 0.0544 &	0.0669 & 0.0222	& 	0.0199 &	0.0400 &	0.0347 & 0.0944 &	0.0439\\
&\ru MTF-GLP-HPM \cite{Alparone2017}	  & 0.0620 & 	0.0722 & 0.0310 &	0.0210 &	0.0508 &	0.0376 & 0.0909 &	0.0411\\
&\ru MTF-GLP-HPM-H \cite{Lolli2017} & 0.1019 &	0.1177 & 0.0270 &	0.0200 &	0.0524 &	0.0401 & 0.0885 &	0.0402\\
&\ru MTF-GLP-HPM-R \cite{Vivone2018} & 0.0497 &	0.0620 & 0.0272 &	0.0210 &	0.0453 &	0.0362 & 0.0928 &	0.0420\\
&\ru MTF-GLP-CBD \cite{Alparone2007}  &	0.0551 &	0.0657 & 0.0222 &	0.0200 &	0.0402 &	0.0346 & 0.0942 &	0.0441\\
&\ru C-MTF-GLP-CBD  \cite{Restaino2017} &	0.0089 &	0.0375 & 0.0224	& 	0.0225 & 	0.0360 &	0.0347 & 0.1129 &	0.0495\\
&\ru MF \cite{Restaino2016}		   	  & 0.0585 &	0.0711 & 0.0421 &	0.0281 &	0.0681 &	0.0441 & 0.0733 &	0.0371\\ \hline
\multirow{3}{*}{VO}&\ru FE-HPM  \cite{Vivone2015a}		  & 0.0579 &	0.0666 & 0.0340 &	0.0210 &	0.0529 &	0.0355 & 0.0784 &	0.0406\\
&\ru SR-D \cite{Vicinanza2015}		  & 0.0225 &	0.0381 & 0.0043	& 	0.0064 &	0.0117 &	0.0190 & 0.1254 &	0.0450\\
&\ru TV	\cite{Palsson2014}		  & 0.0166 &	0.0238 & 0.0225 &	0.0173 &	0.0459 &	0.0373 & 0.0583 &	0.0252\\ \hline
\multirow{4}{*}{ML}&\ru PNN	  \cite{Masi2016}        & 0.0589 &	0.0553 & 0.0740	& 	0.0629 & 	0.1477 &	0.1307 & 0.1815 &	0.0878\\
&\ru PNN-IDX	\cite{Masi2016}	  & 0.0837 &	0.0848 & 0.2671 &	0.0546 &	0.1570 &	0.1005 & 0.3344 &	0.0889\\
&\ru A-PNN	\cite{Scarpa2018a}	  & 0.0527 &	0.0636 & 0.0371 & 	0.0312 &	0.0773 &	0.0607 & 0.0963 &	0.0522\\
&\ru A-PNN-FT \cite{Scarpa2018a}	  & 0.0196 &	0.0203 & 0.0414 &	0.0308 &	0.0815 &	0.0579 & 0.0830 &	0.0489\\
\hline \hline
\end{tabular}
\label{tab:results}
\end{table}

\begin{figure}
\centering
\includegraphics[scale=1]{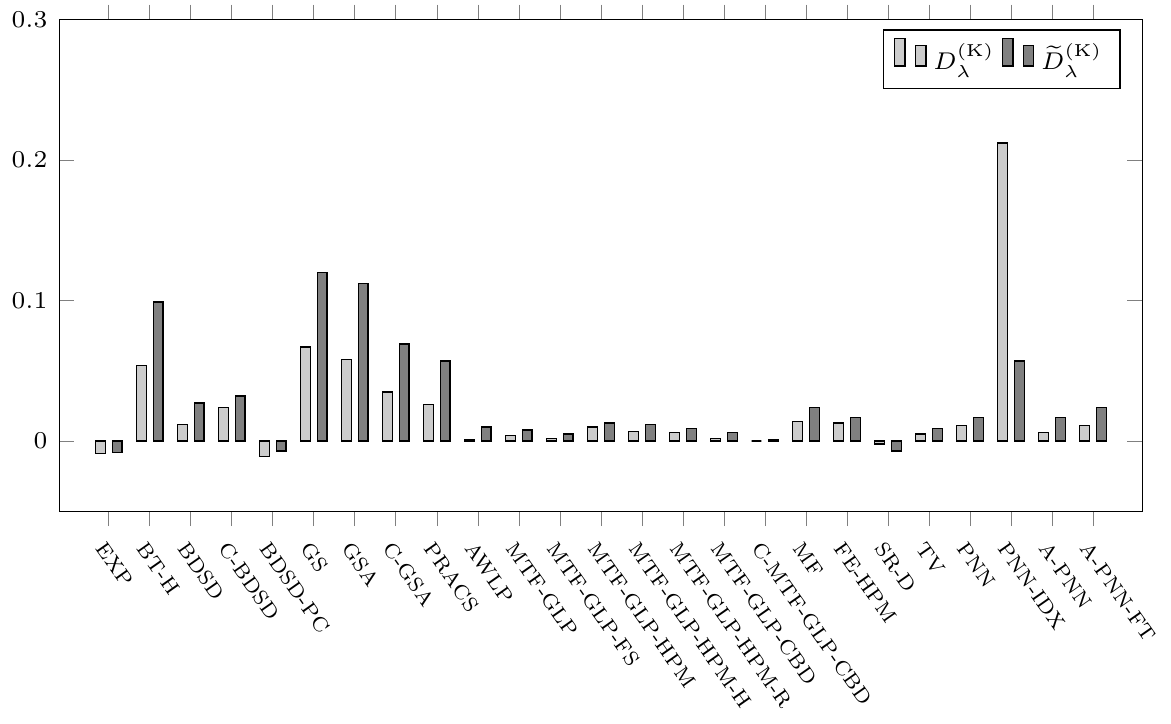}
\caption{Misregistration impact on $\DLK$ and $\DLV$. Bar levels indicate the variation of the indexes 
due to data misragistration.}
\label{fig:mis}
\end{figure}

The impact of bands misregistration on the quality of the fused products has been already recognized in the past \cite{Baronti2011, Jing2012, Xu2014}.
Here we propose an {\em ad hoc} experimental analysis to show the robustness of the proposed 
reprojection indexes to the data mis-registration.
The starting point is a WV3 dataset composed of ten 2048$\times$2048 tiles extracted from a larger image of Adelaide (DigitalGlobe$^\copyright$ sample product).
This dataset presents bands misalignment that we have corrected to produce a companion aligned dataset.
In Tab.~\ref{tab:results} we summarize the average spectral distortion scores for each method for both datasets.
For a handy reading of these numbers, 
in Fig.~\ref{fig:mis} we show the impact of data misregistration on $\DLK$ and $\DLV$ in differential terms.
Each bar indicates the difference between the values of the given indicator computed on aligned and misaligned datasets, respectively.
As it can be clearly noticed, 
traditional CS methods such as BT-H, GS, GSA, C-GSA, PRACS,
that by construction provide fused images 
that are strongly anchored to the PAN geometry,
pay a considerable spectral loss according to $\DLK$ or $\DLV$.
These results are not aligned with our expectations,
as these CS methods are expected to be more robust to misregistration.
The Reader can refer to \cite{Baronti2011} for a theoretical comparison between CS and MRA methods in presence of misregistration,
with the former category being superior to the latter.
 
\begin{figure}
\centering
\includegraphics[scale=1]{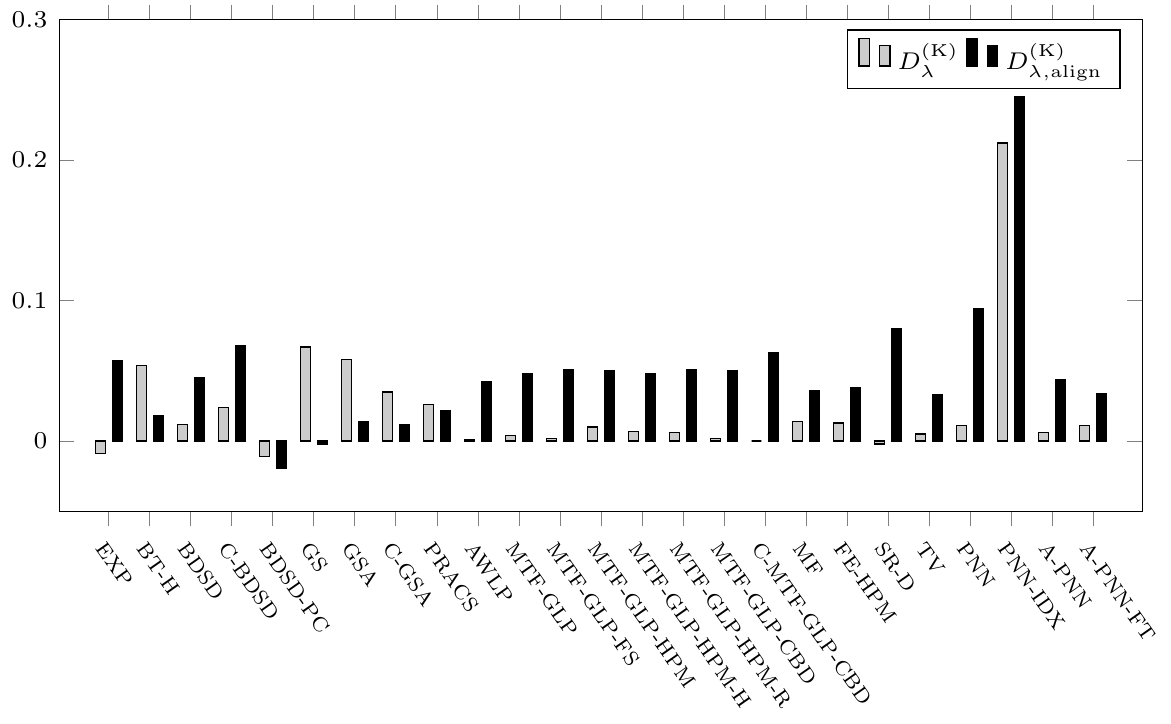}
\caption{Misregistration impact on $\DLK$ and $\DLa$. Bar levels indicate the variation of the indexes 
due to data misragistration.}
\label{fig:mis new}
\end{figure}

In Fig.~\ref{fig:mis new}, instead, it is compared Khan's index $\DLK$ with its variant $\DLa$ with alignment.
Again, bars indicate their variations due to dataset misalignment.
By inspecting the figure it can be noticed that misregistered data 
generally cause an increase of $\DLa$, except for some CS methods,
such as BT-H, GS, GSA, C-GSA and PRACS, 
which tend to preserve the PAN geometry operating an intrinsic alignment. 
On the contrary, 
other methods, notably those belonging to the MRA, VO and ML categories,
that are oriented to the spectral preservation but do not operate alignment,
register an increase of the spectral distortion when the metric takes into account the misalignment.

Concluding, 
the proposed experiment suggests that Khan's index with alignment correction,
that is the complement of the proposed reprojection index R-$Q2^n$, 
is more robust to misalignment or, at least, more consistent with the theoretical expectation \cite{Baronti2011},
than the original Khan's index $\DLK$ or its variant $\DLV$.

\subsection{Reference {\em vs} no-reference indexes cross-checking in the reduced resolution space}
\label{sec:cross-check}
In order to assess the consistency between no-reference and reference-based indicators,
we have designed a set of experiments in the reduced resolution space.
Despite no-reference indexes are conceived to work in the full-resolution framework,
their use on (simulated) reduced resolution datasets allows us to carry out studies 
on the correlation between them and objective error measurements (reference-based indexes) 
thanks to the availability of GTs.
In particular,
for this experiment we have resorted to our WV2 dataset\footnote{For the sake of brevity, 
analogous results obtained on WV3 images will not be presented.} composed of twenty 
2048$\times$2048 images at full-resolution.
These images come from two larger images of Washington (13) and Stockholm (7), respectively,
courtesy samples of DigitalGlobe$^\copyright$.
Each such tiles was resized to 512$\times$512 pixels using the usual Wald's downgrading protocol.
This dataset was already well coregistered, 
therefore we have proceeded to create a misregistered counterpart 
operating a simple modification of the downgrading process:
a 1-pixel shift (in both directions) has been introduced in the decimation (after LPF) of the MS bands.

\begin{figure}
\centering
\includegraphics[scale=1]{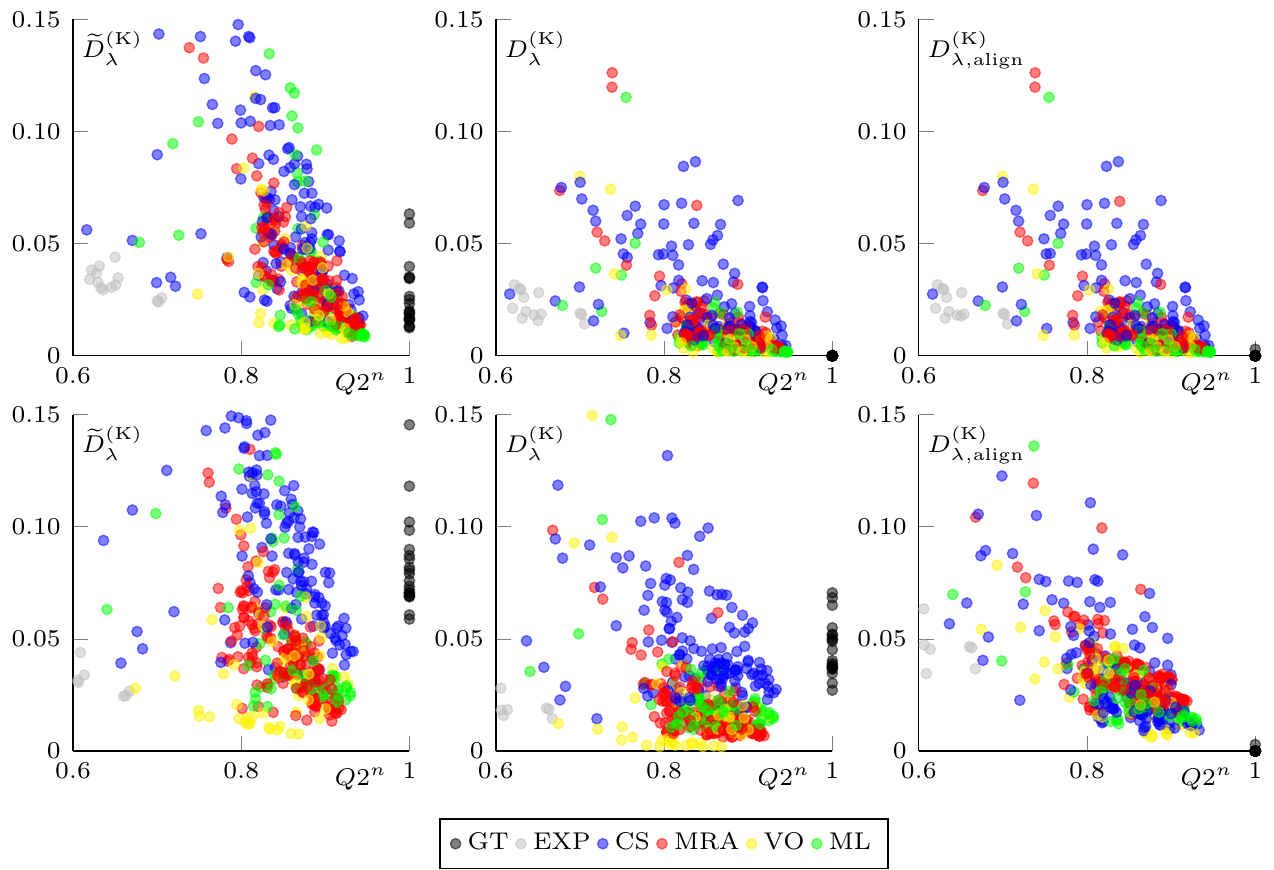}
\caption{From left to right $\DLV$, $\DLK$ and $\DLa = 1-$R-$Q2^n$ {\em vs} $Q2^n$ on the WV2 RR dataset.
Both the ``aligned'' (top) and ``mis-aligned'' (bottom) dataset cases are shown.
Each marker is associated to a single pansharpening result on a $512\times512$ tile.
Black markers correspond to the ground-truth. Light gray ones (EXP) correspond to a simple interpolation.
The remaining clusters correspond to the different algorithms grouped by categories.}
\label{fig:scatter-misreg}
\end{figure}

Then, 
25 pansharpening algorithms provided by the toolbox \cite{Vivone2020} were run on each RR sample image
generating 500 results, for each of the two datasets (registered and not),
for which all the indexes of interest have been computed.
Eventually, we have obtained hundreds of points in a multi-dimensional evaluation space, which enable plenty of analyses,
with some dimensions corresponding to the reference-based indexes (SAM, ERGAS and $Q2^n$, SSIM, CMSC),
and the remaining ones associated to the proposed indexes (R-SAM, R-ERGAS, R-$Q2^n=1-\DLa$ , $\DR$) 
and to other state-of-the-art no-reference indexes ($\DL$, $\DLK$, $\DLV$, $\DS$, QLR$_1$, QLR$_2$).
In particular, 
by construction we expect to observe a good correlation between Khan's index variants and $Q2^n$.
Therefore we start from the scatter plots shown in Fig.~\ref{fig:scatter-misreg} with such variants, in turn, {\em vs} $Q2^n$.
\begin{figure}
\centering
\includegraphics[scale=1]{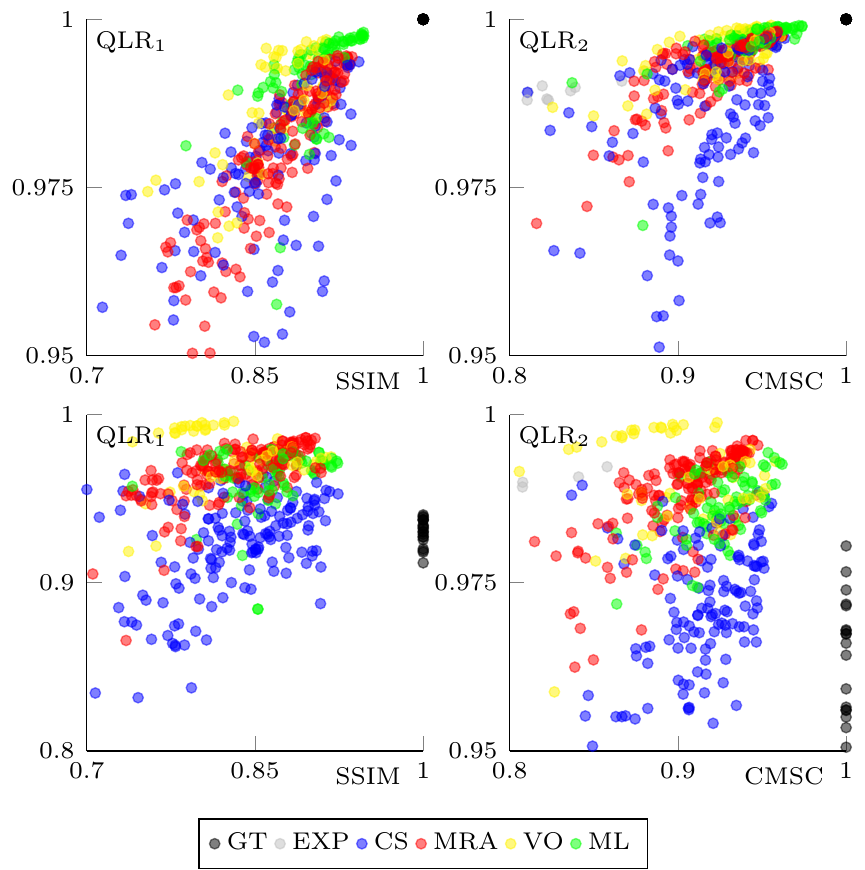}
\caption{From left to right QLR$_1$ {\em vs} SSIM and QLR$_2$ {\em vs} CMSC on the WorldView-2 RR dataset.
Both the ``aligned'' (top) and ``mis-aligned'' (bottom) dataset cases are shown.
Each marker is associated to a single pansharpening result on a $512\times512$ tile.
Black markers correspond to the ground-truth. Light gray ones (EXP) correspond to a simple interpolation.
The remaining clusters correspond to the different algorithms grouped by categories.}
\label{fig:scatter-competitor}
\end{figure}
Both the ``aligned'' (top) and the ``mis-aligned'' (bottom) datasets are considered.
For the aligned case, 
by visual inspection we can appreciate that $\DLK$ correlates better than $\DLV$ with $Q2^n$.
Besides, $\DLa$ behaves just like $\DLK$ because no alignment is operated.
For both $\DLK$ and $\DLa$, the GTs (black dots), for which $Q2^n$ is always 1, 
get the ideal value (0) of spectral distortion.\footnote{Rarely, 
the alignment process (based on the maximization of the correlation) 
embedded in the computation of $\DLa$ fails for some spectral band giving raise to non-zero results for the GT even if the test image is aligned.}
Besides,
$\DLV$ is not minimized by the GT,
coherently with its definition (Eq.~\ref{eq:dl}) based on the comparison between the smoothed GT and the upscaled MS.
It is also worth to notice that the correlation degree between $Q2^n$ and the different variants of spectral distortion
grows when assessed by category (by colors), 
supporting the idea that no-reference indexes are more reliable for intra-class assessment \cite{Vivone2020}. 
Moving to mis-aligned dataset (bottom scatters),
it can be clearly recognized a bias for both $\DLV$ and $\DLK$, 
which do not occurs for $\DLa$,
represented by the shift of the GT scores which no longer get the ideal value.
More in general, the distortion scores in terms of $\DLV$ and $\DLK$ register a degradation,
with CS methods (blue) much more penalized than others.
This last observation gives a further support the interpretation provided above about the results of Fig.~\ref{fig:mis} and Fig.~\ref{fig:mis new}.
A similar behavior is also registered other state-of-the-art indexes. 
For example, in Fig.\ref{fig:scatter-competitor} the score scatters in the QLR$_1$-SSIM (left) and QLR$_2$-CMSC (right) planes, 
with registered (top) and misaligned (bottom) datasets.

\begin{figure}
\centering
\includegraphics[scale=1]{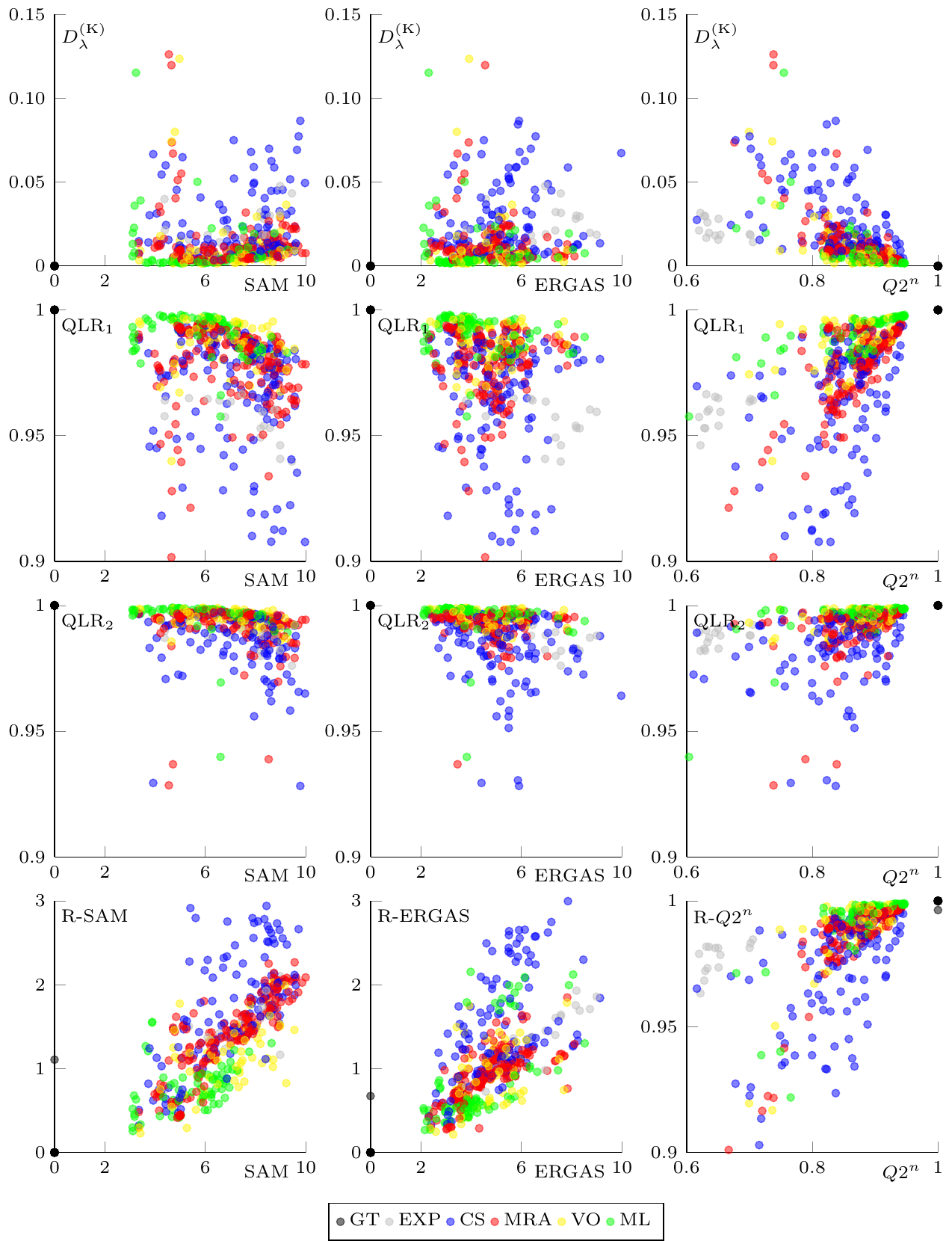}
\caption{No-reference {\em vs} reference-based indexes scores on the aligned WV2 RR dataset.
From left to right, SAM, ERGAS and $Q2^n$ are selected as reference-based indexes ($x$-axis).
From top to bottom, $\DLK$, QLR$_1$, QLR$_2$ and R-$\mathcal{M}$ are chosen as no-reference indexes ($y$-axis).}
\label{fig:scatter-all}
\end{figure}

Let us now leave the registration problem out considering aligned datasets only 
and looking at the relationship between the spectral consistency indexes and the three most used reference-based indexes, {\em i.e.}, SAM, ERGAS and $Q2^n$,
with the help of Fig.~\ref{fig:scatter-all}.
From top to bottom we can see how different compared spectral consistency indexes agrees with the three (column-wise distributed) reference-based indexes.
From the top row scatters it clearly appears that $\DLK$ agrees with SAM and ERGAS much lesser than with $Q2^n$.
This because $\DLK$ is based on the same $Q2^n$ index but also because SAM and ERGAS encode a different concept of quality.
Similar considerations apply for QLR$_1$ and QLR$_2$ 
who, as well as $Q2^n$, are both based on the comparison among local statistics.
For these reasons, 
we believe that in addition to R-$Q2^n$ it makes sense to also provide R-SAM and R-ERGAS for a more comprehensive evaluation of pansharpening methods.
On the bottom row, we show all three ($\mathcal{M}$, R-$\mathcal{M}$) scatter plots,
which speak in favor of a good level of agreement between each objective index $\mathcal{M}$ and the reprojected counterpart R-$\mathcal{M}$. 

Beside the qualitative interpretation of the score scatters, 
we can quantify the level of agreement among the reference-based indexes and the compared no-reference indexes in terms of correlation coefficients.
These are shown in Tab.~\ref{tab:cc} in the usual matrix form, for both the aligned (a) and misaligned (b) dateset.
As expected the reprojected indexes show a relatively high correlation with the non-reprojected counterpart, both for aligned and misaligned datasets.
$\DLK$ and $\DLV$ also correlate well with $Q2^n$, but only in the aligned case.
Likewise, QLR$_1$ and QLR$_2$ also correlate well with different reference-based indexes in ideal conditions but register a drop on misaligned data.
It is also worth to remark that GT scores have been discarded to not penalize excessively the competitors on the misaligned dataset 
(see GT score distribution in Fig.~\ref{fig:scatter-misreg}-\ref{fig:scatter-competitor})
\begin{table}
\centering
\caption{Correlation coefficient among different indexes (some have been complemented for a uniform polarization) obtained on the 
aligned (a) and misaligned (b) RR WV2 datasets. Outlier and GT scores have been discarded.}
\begin{tabular}{c}
\includegraphics[scale=0.7]{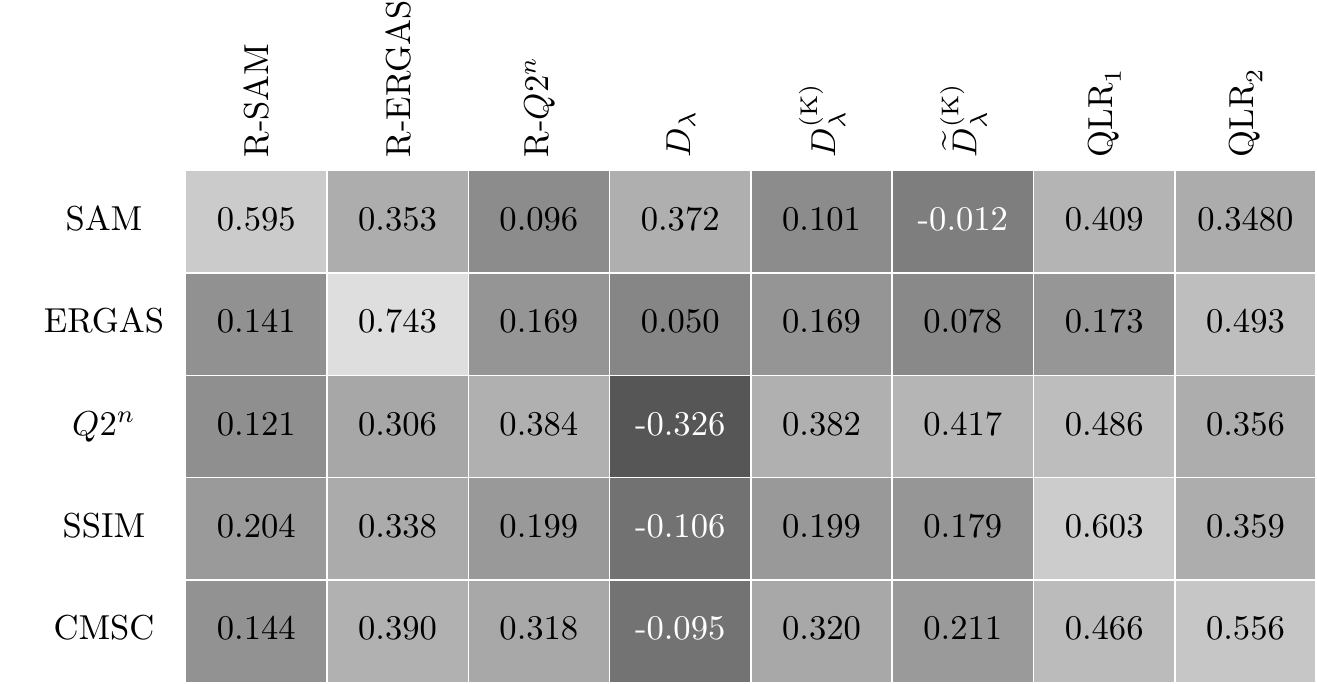} \\
(a) \\[3mm]
\includegraphics[scale=0.7]{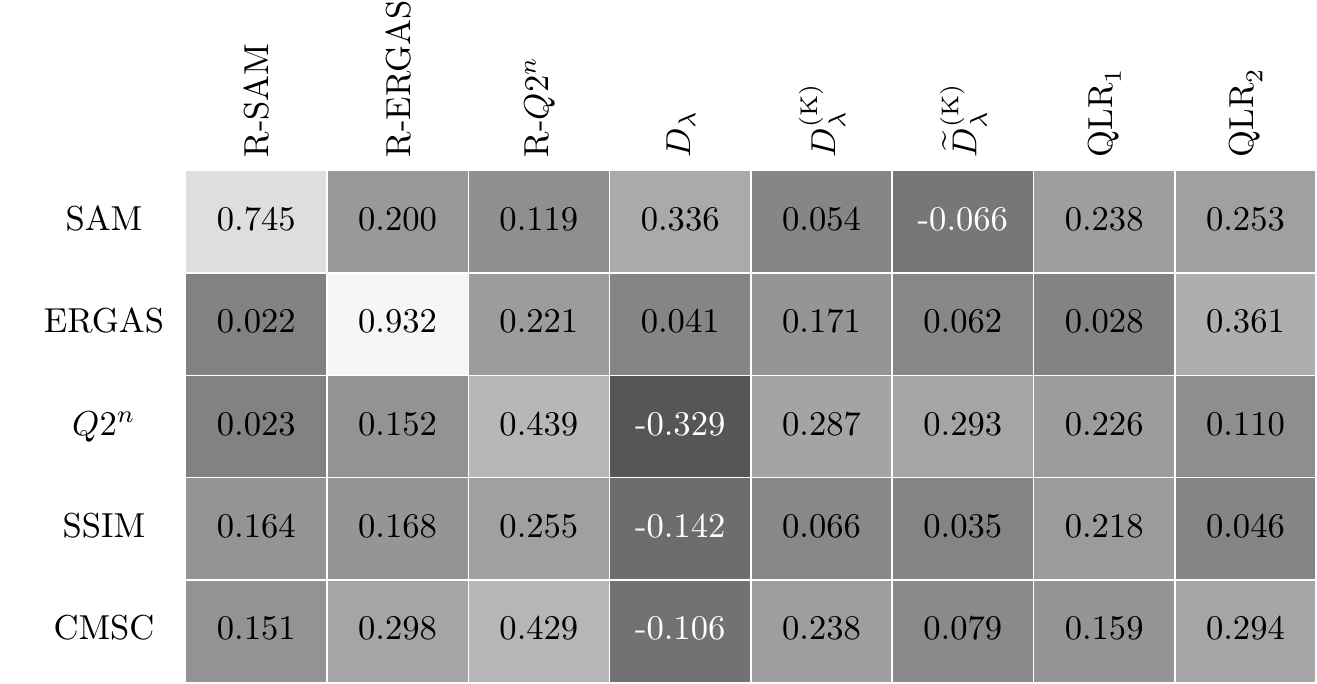} \\
(b)
\end{tabular}
\label{tab:cc}
\end{table}

\subsection{A qualitative assessment of the proposed spatial distortion index}
While objective synthesis quality indexes such as $Q2^n$ and ERGAS 
account for both spectral and spatial inaccuracies, their reprojected versions clearly 
limit their scope to the spectral component as they are computed on low-pass versions of the fused products
ignoring image ``details''. 
It is therefore necessary to complement these indexes with some spatial quality indicator. 

The assessment of the spatial quality of a pansharpened image in the full-resolution framework
is very difficult and somewhat controversial, 
lacking a degradation model to describe the relationship between the full-resolution spatial-spectral datacube (fused image)
and the panchromatic image. 
In fact,
while the spectral degradation can be reasonably modeled through the sensor MTF, allowing for a spectral consistency check,
the spatial degradation cannot be modeled through a simple global weighted band average \cite{Thomas2008}.
Indeed,
in addition to the obvious spatial dependency of the spectral mixing process that provides the PAN image,
there is also a mismatch between the PAN spectral coverage and the MS coverage \cite{Thomas2008}.
This means that there could be details ``seen'' by the PAN but not by the virtual full-resolution MS counterpart, and {\em vice versa}.
This is the origin of an ill-posedness of the pansharpening problem, mostly residing in the spatial reconstruction,
which makes the spatial consistency assessment subjective to some extent.
On the basis of this observation, 
we provide here an interpretation of some sample results through visual inspection,
comparing the proposed index $\DR$ with the state-of-the-art spatial distortion index $\DS$.
Of course, 
given the subjective nature of this kind of analysis, 
we will focus on clearly visible phenomena, leaving to the Reader the final say based on his/her visual perception of more subtle patterns, if any.
In particular, we propose two experiments: an ``horizontal'' comparison among the pansharpening toolbox methods
and a ``vertical'' comparison where a single machine learning method is optimized using the proposed $\DR$ (varying the related scale parameter $\sigma$) 
jointly with a term controlling the spectral consistency.

\begin{figure}
\centering
\begin{tabular}{c}
\includegraphics[scale=0.9]{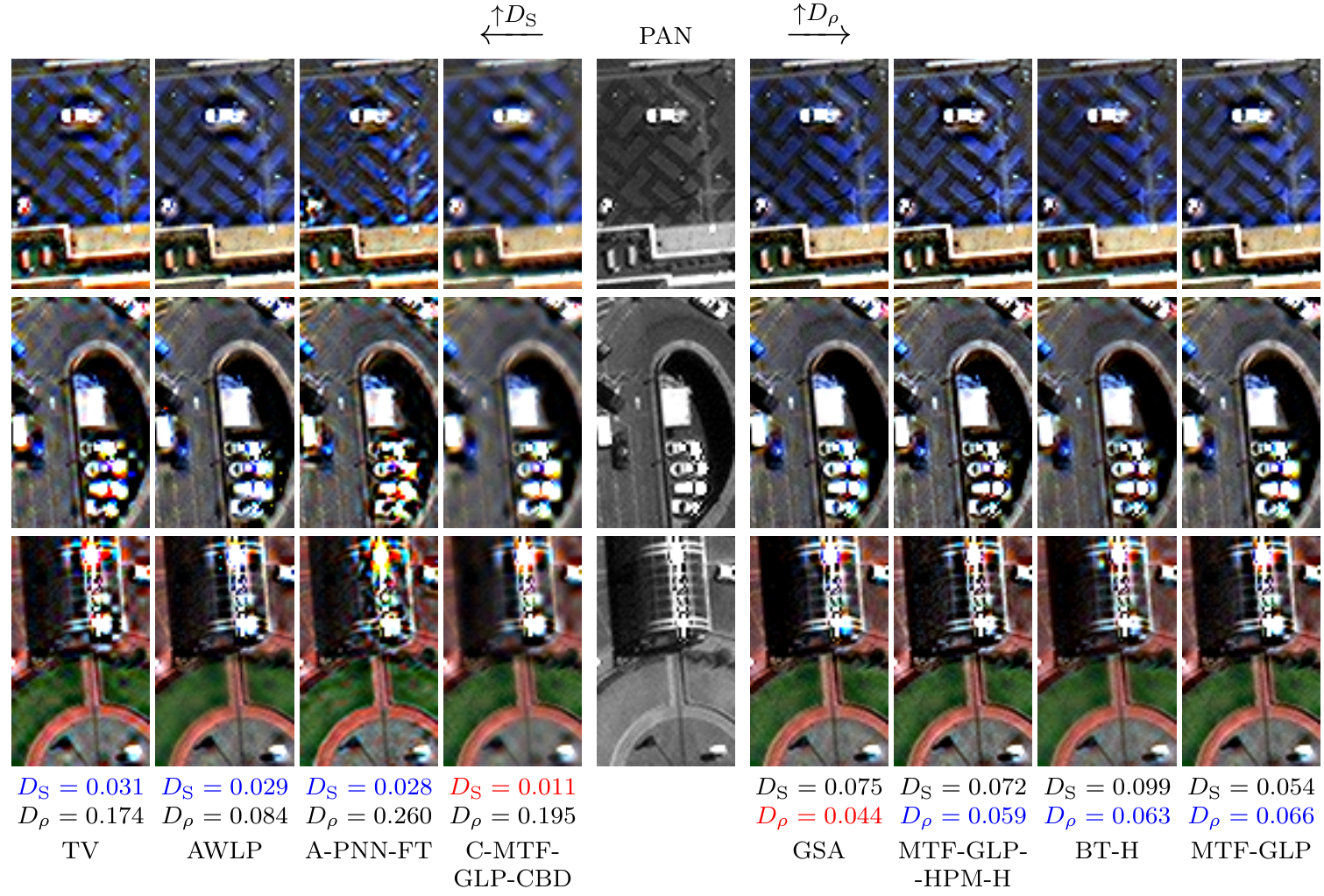}\\[3mm]
\includegraphics[scale=0.9]{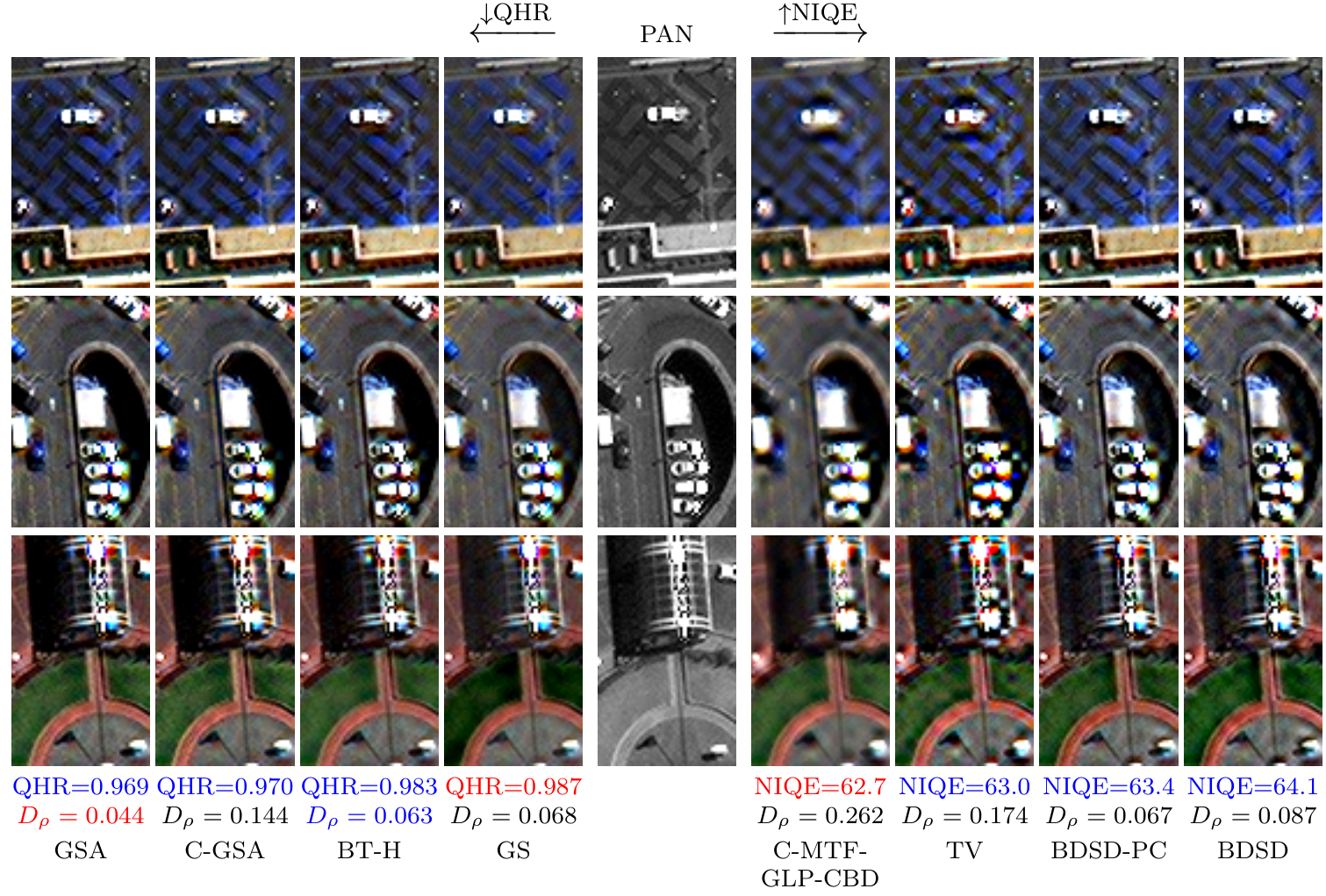}
\end{tabular}
\caption{Pansharpening results on crops from a full-resolution WV3 Adelaide tile.
The reference PAN is in the middle column.
In the top half the four best results in terms of $\DS$ and $\DR$ are on its left and right, respectively, with best scores (in red) closer to the reference.
In the bottom half are shown the top results according to QHR (leftward) and NIQE (rightward).
Best scores are highlighted in red; second, third and fourth bests are in blue.}
\label{fig:pansharp}
\end{figure}

Let us start with the first experiment. 
Fig.~\ref{fig:pansharp} shows some FR pansharpening results for crops extracted from a single tile of the WV3 Adelaide image.\footnote{For 
visualization purposes, hereinafter all displayed crops extracted from multispectral images are obtained combining the red, green and blue channels (see Tab.~\ref{tab:sensors}).}
The PAN component, used as a reference, is shown in the middle of different groups of results.
In the top half part of the figure, the top-4 solutions according to $\DS$ (left) and the top-4 according to $\DR$ (right) are shown with $\DS$ and $\DR$ computed on the whole tile.
Similarly, in the bottom half part, the top-4 results according to QHR and NIQE are shown with related scores.
It appears clearly that images selected according to the $\DR$ index
ensure a better agreement with the reference in terms of spatial layout and, more in general, a better quality,
with sharp contours, accurate textures, and the lack of annoying patterns such as those present in some top-$\DS$ and top-NIQE images.
It is interesting to observe that, 
among the top-4 $\DS$ results, 
the most convincing one from visual inspection
corresponds also to the relative lowest $\DR$ (AWLP).
It is also worth to observe a certain degree of agreement between $\DR$ and QHR.
Similar phenomena are observed on all other tiles.

In the previous experiment
we have set the scale parameter $\sigma=R=4$ according to the above discussed theoretical motivations (Section~\ref{sec:rho}).
To gain insight into the effectiveness of this choice we have designed an additional {\em ad hoc} experiment 
to provide pansharpening results with controlled spectral and spatial qualities,
with the latter quantified through $\DR$, configurable using different scale settings.
This is achieved by leveraging on a CNN model working in target-adaptive modality \cite{Ciotola2022} and using 
a combination of the wanted consistency measures as a loss.
In this context,
the choice of the specific CNN model is not critical as we work in adaptive modality, 
by running unsupervised tuning iterations on each test image until the loss terms reach the desired level (overfitting).
In particular, the optimized loss is defined as

\begin{equation}
\mathcal{L} = \max(\|\hMda^{\rm lp} - M \|_1, \mathcal{L}_\lambda^*)
					+ \beta \max(\DR^{(\sigma)}, \DR^*),
\label{eq:loss}
\end{equation}
where $\mathcal{L}_\lambda^*$ and $\DR^*$ are two prefixed target levels for the spectral ($\|\hMda^{\rm lp} - M \|_1$)
and spatial ($\DR^{(\sigma)}$) quality indicators, respectively.
In practice, 
the network parameters are pushed to overfit the test image so that the corresponding loss reaches the target quality $(\mathcal{L}_\lambda^*, \DR^*)$.
These two threshold values could be ideally set to zero, but this would lead to extremely long tuning 
and may also generate instability because of a conflicting interaction between the two loss terms. 
In particular, we have set $\mathcal{L}_\lambda^*=0.015$, quite a low value considering the dynamic range of the pixels values,
and $\DR^*=0.1$, which is also quite low according to our experiments.
By doing so, each test requires a different number of tuning iterations. We have therefore oversized the number (experimentally set to 5000) of iterations to ensure convergence for all.
This process is repeated for several choices of the scale parameter $\sigma$ ranging from 2 to 64.
In Fig.~\ref{fig:sigma}, 
we show some crops from the WV3 dataset.
These sample results reflect a general behavior that we have observed in a wide range of experiments,
that is a relatively good response in the range of $\sigma \in [4,8]$. For smaller ($\sigma = 2$) or larger ($\sigma>8$) scale values noisy patterns arise.
These are particularly noticeable on roofs and roads for the selected samples.
The above observations provide an experimental validation of the choice $\sigma=R=4$
proposed in this work on the basis of the theoretical motivations discussed in Section~\ref{sec:rho}. 

\newcommand{\pathSigma}{./Figures/sigma_effects/}
\newcommand{\imSigma}[1]{\includegraphics[width=1.65cm]{\pathSigma #1.png}}
\begin{figure}
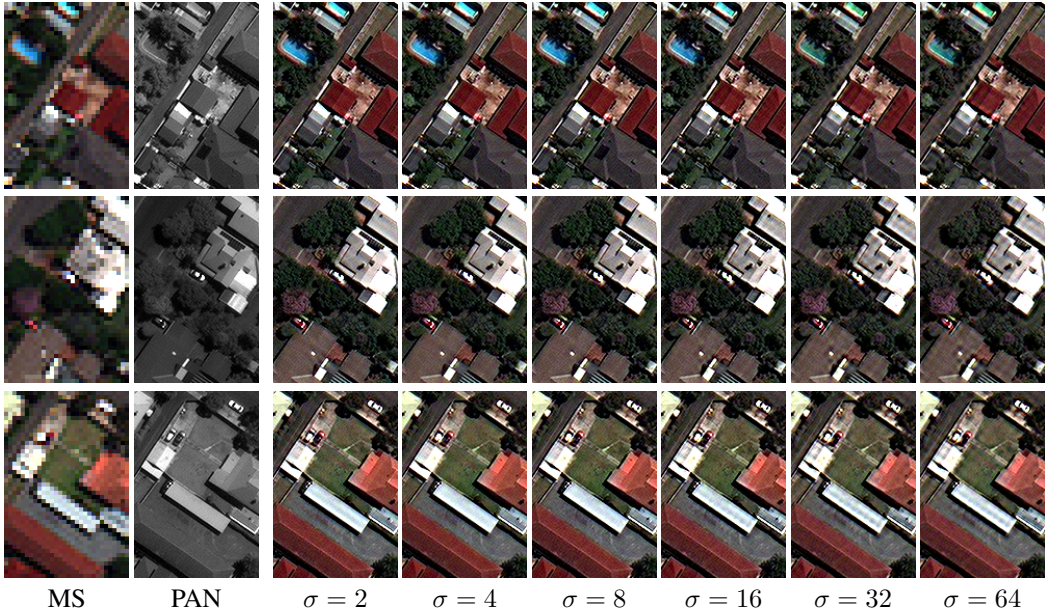

\centering
\setlength{\tabcolsep}{1pt}
\begin{tabular}{cc@{\rule{2mm}{0mm}}cccccc}
\imSigma{MS_1} & \imSigma{PAN_1} & \imSigma{scale_2_1} & \imSigma{scale_4_1} & \imSigma{scale_8_1} & \imSigma{scale_16_1} & \imSigma{scale_32_1} & \imSigma{scale_64_1} \\ 
\imSigma{MS_2} & \imSigma{PAN_2} & \imSigma{scale_2_2} & \imSigma{scale_4_2} & \imSigma{scale_8_2} & \imSigma{scale_16_2} & \imSigma{scale_32_2} & \imSigma{scale_64_2} \\ 
\imSigma{MS_3} & \imSigma{PAN_3} & \imSigma{scale_2_3} & \imSigma{scale_4_3} & \imSigma{scale_8_3} & \imSigma{scale_16_3} & \imSigma{scale_32_3} & \imSigma{scale_64_3}  \\ 
MS & PAN & $\sigma = 2$ &  $\sigma = 4$ & $\sigma = 8$ &  $\sigma = 16$ &  $\sigma = 32$ &  $\sigma = 64$
\end{tabular}
\caption{Pansharpening results with controlled spectral ($\ell_1$-norm) and spatial ($\DR$) distortions. 
From left to right: MS and PAN, followed by the outputs optimized using $\sigma\in\{2,4,8,16,32, 64\}$.}
\label{fig:sigma}
\end{figure}

The present experiment gives us the opportunity to also make some
considerations about the computational load related to the proposed indexes. 
In general, in the context of quality assessment, 
the computation time is not a critical issue as it is carried out offline. 
However, with the advent of deep learning,
researchers started to use such quality indexes as loss functions for training purposes.
In this regard, 
it is worth to distinguish pixel-based indexes ({\em e.g.}, SAM, ERGAS, $\ell_\alpha$-norms),
that are relatively light to compute, from other indexes that involve local statistics computed 
at a certain scale for each location ({\em e.g.}, $Q2^n$, $\DR$, $\DLK$).
When using the latter as loss the training may slowdown.
In the particular case of $\DR$, that has been included in the loss function to finetune the sample CNN employed in the experiment,
we have actually registered a moderate impact on the training time.
In fact, 
using a training batch composed of a single 2048$\times$2048 image,
the time consumption per iteration, without or with the additional $\DR$ loss component (second term in Eq.~\ref{eq:loss}),
shifts from 1.27s to 1.6s on a NVIDIA P6000 GPU.

\subsection{Comparative results}
\begin{figure}
\centering
\includegraphics[scale=1]{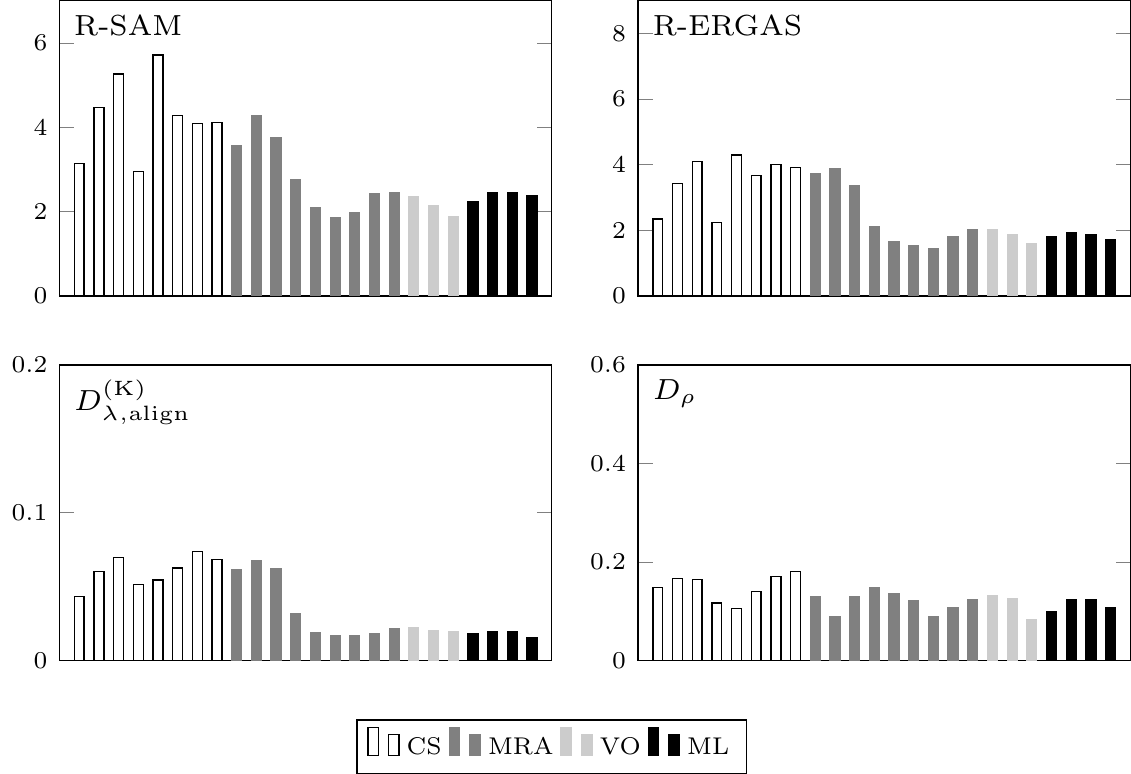}
\caption{Full-resolution assessment on the WV3 dataset.
R-$Q2^n$ is replaced by its complement $\DLa$ for the sake of readability (0 is the ideal value for all). For details on methods and related categories (CS, MRA, VO, ML) refer to Tab.~\ref{tab:results}.}
\label{fig:results}
\end{figure}

\begin{figure}
\centering
\includegraphics[scale=1]{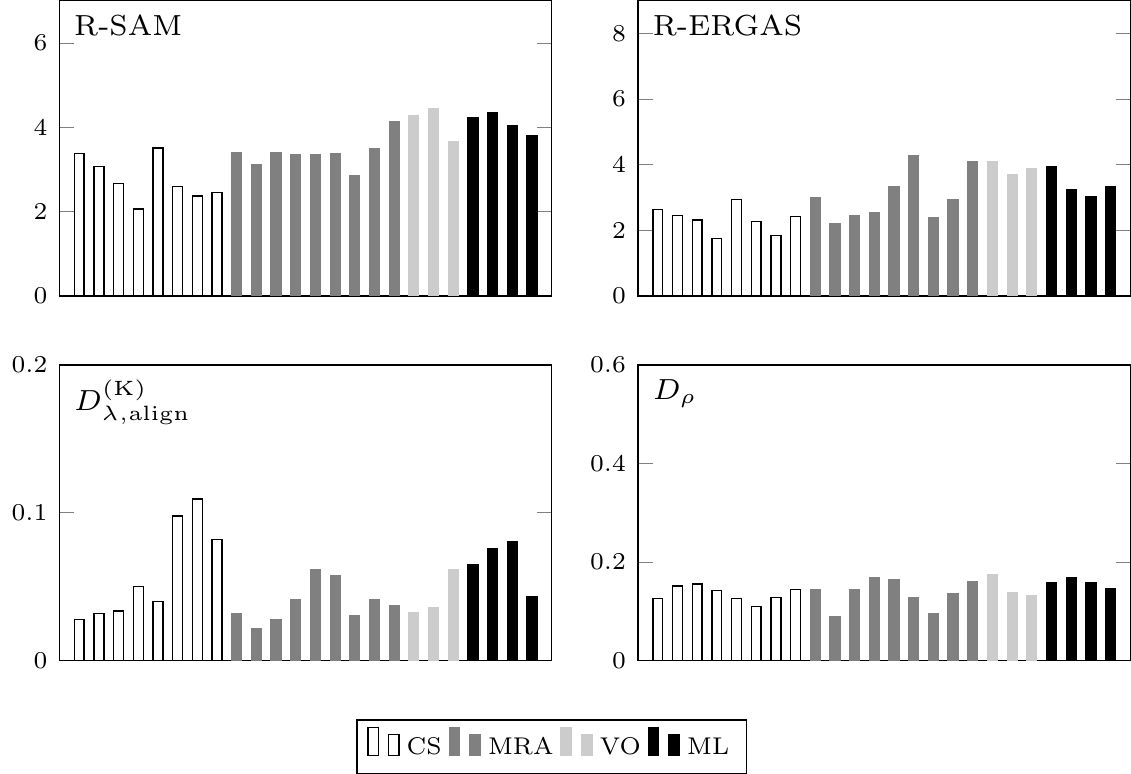}
\caption{Full-resolution assessment on the WV2 dataset.
R-$Q2^n$ is replaced by its complement $\DLa$ for the sake of readability (0 is the ideal value for all). For details on methods and related categories (CS, MRA, VO, ML) refer to Tab.~\ref{tab:results}.}
\label{fig:resultsWV2}
\end{figure}

To conclude,
we present here an overall comparison among the available pansharpening methods provided by the toolbox \cite{Vivone2020}, 
using the proposed quality indicators and supported by the visual inspection of some sample results.
In Fig.~\ref{fig:results} and Fig.~\ref{fig:resultsWV2} we gather the average numerical results obtained on our WV3 and WV2 datasets, respectively.
As expected, 
the numerical results clearly show a good level of agreement among the three reprojection error indexes,
all essentially linked to the spectral consistency. 
However, some exceptions can be observed, particularly for the WV2 dataset (Fig.~\ref{fig:resultsWV2})
where some CS methods register a performance loss in terms of $\DLa$ not observed on R-SAM and R-ERGAS.
We also recognize a good agreement with some literature findings, 
particularly on WV3 (Fig.~\ref{fig:results}), 
such as the spectral accuracy gap between MRA and CS methods \cite{Vivone2015}
or the competitiveness of some MRA, VO and ML solutions.
Moreover, for the ML solutions, 
it is worth to notice a performance gap moving from one dataset (WV3) to another (WV2).
Indeed, such variability is not unusual for data-driven approaches that can suffer generalization limits 
when the training dataset is not sufficiently representative.
In this particular situation,
it could likely be the case that the WV2 test images used are too different from the images used for the training of the involved ML methods.

To gain insight into the effectiveness of the proposed indexes let us give a look to some sample results
for a complementary subjective analysis. 
In Fig.~\ref{fig:WV3FRcrops} we show some clips from a single WV3 tile of the Adelaide dataset.
On the leftmost column are gathered the PAN and MS input components on two consecutive rows.
Then, the corresponding pansharpening results of the best performing solutions according to R-$Q2^n$
are shown in decreasing order from left to right, next to the PAN.
Next to the MS, we also show the ``reprojection'' error map, 
that is the difference between the input MS and the reprojection (LPF plus decimation) 
of the output. 
On the bottom lines are gathered the R-$Q2^n$ and $\DR$ scores.
Therefore, the spectral quality of the results decreases left to right. 
This can also be appreciated through the inspection of the reprojection error maps.
Besides,
the best methods from the spatial perspective ($\DR$) follow a different ordering.
This partially reflects a tradeoff between spectral and spatial features.
Particularly interesting is the case of the method TV which scores first in terms of R-$Q2^n$
(actually it is the best according to R-SAM and R-ERGAS, as well)
but shows the worst $\DR$ value, 0.215, 
corresponding to a 78.5\% average correlation between the PAN and the pansharpened bands.
The impact of this low correlation level is clearly visible in the pansharpening results of TV
which show underlying noisy patterns.
Such patterns disappear with the reprojection explaining why the reprojection indexes do not worsen.
The other methods achieve much lower values of $\DR$, ranging from 0.084 to 0.115
(around 90\% of correlation),
to which correspond to a higher coherence between the spatial features of the results and the PAN,
easily appreciable through visual inspection.
For completeness, 
in Fig.~\ref{fig:WV2FRcrops} we show analogous results for a WV2 tile,
which basically confirms the same considerations made above for WV3.

\newcommand{\pathW}{./Figures/DetailsWV3/}
\newcommand{\imW}[1]{\includegraphics[width=1.6cm]{\pathW #1.png}}
\begin{figure}
\centering
\scriptsize
\setlength{\tabcolsep}{1pt}
\begin{tabular}{c@{\rule{2mm}{0mm}}ccccccc}
\imW{PAN_1} & \imW{TV_1}        & \imW{MF_1}         & \imW{MTF-GLP-HPM_1}        & \imW{AWLP_1}           & \imW{MTF-GLP-HPM-H_1}           & \imW{FE-HPM_1}         & \imW{MTF-GLP_1}    \\
PAN & TV & MF & ...-GLP-HPM & AWLP & ...-HPM-H & FE-HPM & MTF-GLP \\[1mm]
\imW{MS_1}  &  \imW{DT_TV_1} & \imW{DT_MF_1} & \imW{DT_MTF-GLP-HPM_1} & \imW{DT_AWLP_1}    & \imW{DT_MTF-GLP-HPM-H_1}    & \imW{DT_FE-HPM_1} & \imW{DT_MTF-GLP_1}    \\
MS &  \multicolumn{7}{c}{\footnotesize Reprojection error maps: defference between the MS and the reprojected pansharpening} \\[3mm]
\imW{PAN_2} & \imW{TV_2}        & \imW{MF_2}         & \imW{MTF-GLP-HPM_2}        & \imW{AWLP_2}           & \imW{MTF-GLP-HPM-H_2}           & \imW{FE-HPM_2}         & \imW{MTF-GLP_2}    \\
PAN & TV & MF & ...-GLP-HPM & AWLP & ...-HPM-H & FE-HPM & MTF-GLP \\[1mm]
\imW{MS_2}  &  \imW{DT_TV_2} & \imW{DT_MF_2} & \imW{DT_MTF-GLP-HPM_2} & \imW{DT_AWLP_2}    & \imW{DT_MTF-GLP-HPM-H_2}    & \imW{DT_FE-HPM_2} & \imW{DT_MTF-GLP_2}    \\
MS &  \multicolumn{7}{c}{\footnotesize Reprojection error maps: defference between the MS and the reprojected pansharpening} \\[3mm]
\imW{PAN_3} & \imW{TV_3}        & \imW{MF_3}         & \imW{MTF-GLP-HPM_3}        & \imW{AWLP_3}           & \imW{MTF-GLP-HPM-H_3}           & \imW{FE-HPM_3}         & \imW{MTF-GLP_3}    \\
PAN & TV & MF & ...-GLP-HPM & AWLP & ...-HPM-H & FE-HPM & MTF-GLP \\[1mm]
\imW{MS_3}  &  \imW{DT_TV_3} & \imW{DT_MF_3} & \imW{DT_MTF-GLP-HPM_3} & \imW{DT_AWLP_3}    & \imW{DT_MTF-GLP-HPM-H_3}    & \imW{DT_FE-HPM_3} & \imW{DT_MTF-GLP_3}    \\
MS &  \multicolumn{7}{c}{\footnotesize Reprojection error maps: defference between the MS and the reprojected pansharpening} \\[3mm]
& R-$Q2^n$={\bf 0.950} & R-$Q2^n$=\underline{0.945} & R-$Q2^n$=0.939 & R-$Q2^n$=0.937 & R-$Q2^n$=0.937 & R-$Q2^n$=0.937 & R-$Q2^n$=0.936 \\
& $\DR$=0.215 & $\DR$=0.115 & $\DR$=\underline{0.090} & $\DR$=0.105 & $\DR$=0.096 & $\DR$=0.114 & $\DR$={\bf 0.084} \\
\end{tabular}
\caption{Sample crops from a WV3 sample image (Adelaide) and related full-resolution pansharpening. 
For each crop, in addition to the PAN and MS components (left-most column),
the best performing solutions according to R-$Q2^n$ are shown from left to right in decreasing order, next to the PAN.
For each method, both the R-$Q2^n$ and $\DR$ scores are reported on the bottom lines of the figure.
Moreover, next to the MS, for each pansharpening result the corresponding reprojection error map is shown.}
\label{fig:WV3FRcrops}
\end{figure}

\newcommand{\pathWV}{./Figures/DetailsWV2/}
\newcommand{\imWV}[1]{\includegraphics[width=1.6cm]{\pathWV #1.png}}
\begin{figure}
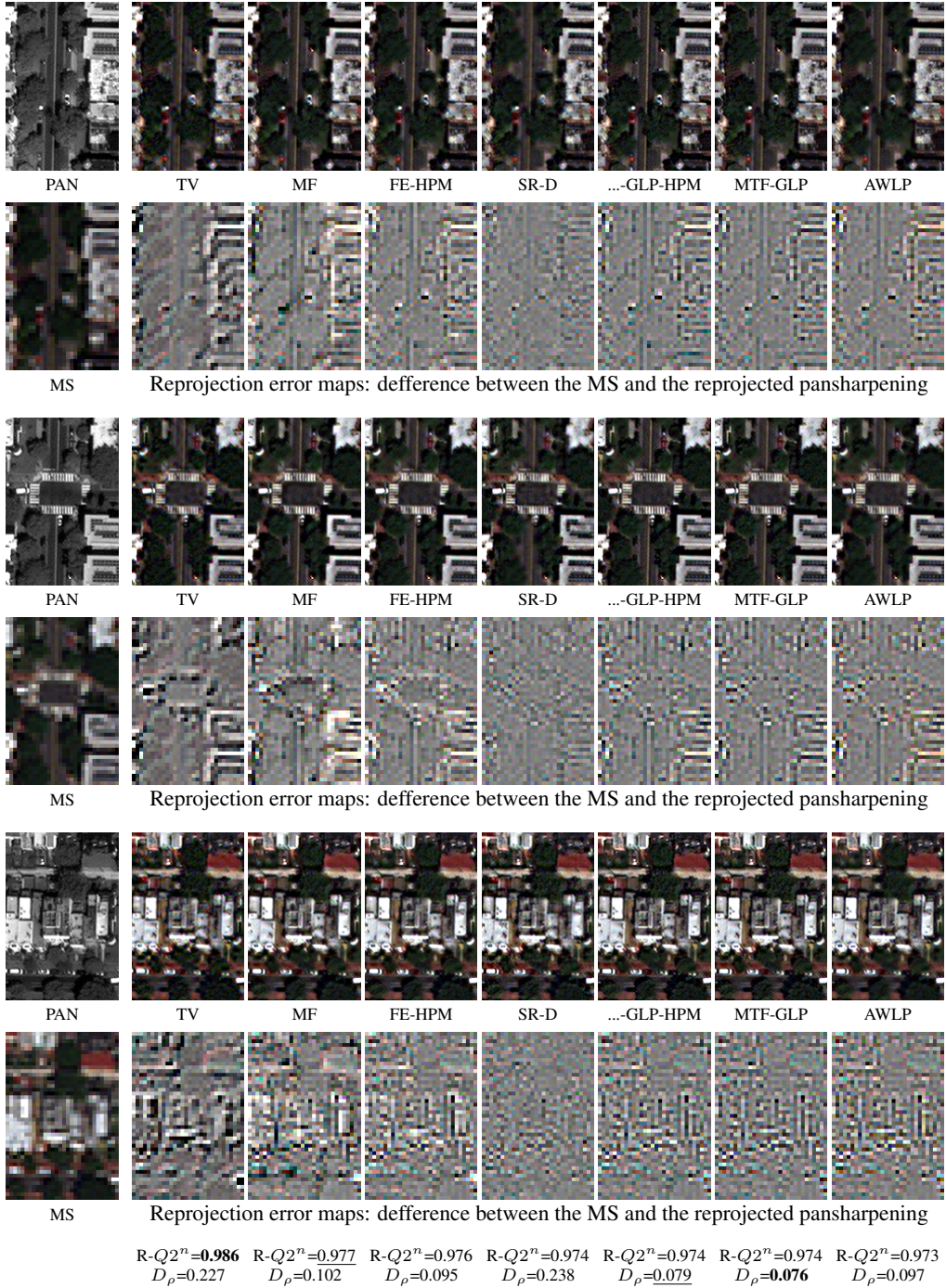

\centering
\scriptsize
\setlength{\tabcolsep}{1pt}
\begin{tabular}{c@{\rule{2mm}{0mm}}ccccccc}
\imWV{PAN_1} & \imWV{TV_1}        & \imWV{MF_1}         & \imWV{FE-HPM_1}        & \imWV{SR-D_1}           & \imWV{MTF-GLP-HPM_1}           & \imWV{MTF-GLP_1}         & \imWV{AWLP_1}    \\
PAN & TV & MF & FE-HPM & SR-D & ...-GLP-HPM & MTF-GLP & AWLP \\[1mm]
\imWV{MS_1}  & \imWV{DT_TV_1}     & \imWV{DT_MF_1}      & \imWV{DT_FE-HPM_1}     & \imWV{DT_SR-D_1}        & \imWV{DT_MTF-GLP-HPM_1}        & \imWV{DT_MTF-GLP_1}       & \imWV{DT_AWLP_1} \\
MS &  \multicolumn{7}{c}{\footnotesize Reprojection error maps: defference between the MS and the reprojected pansharpening} \\[3mm]
\imWV{PAN_2} & \imWV{TV_2}        & \imWV{MF_2}         & \imWV{FE-HPM_2}        & \imWV{SR-D_2}           & \imWV{MTF-GLP-HPM_2}           & \imWV{MTF-GLP_2}         & \imWV{AWLP_2}    \\
PAN & TV & MF & FE-HPM & SR-D & ...-GLP-HPM & MTF-GLP & AWLP \\[1mm]
\imWV{MS_2}  & \imWV{DT_TV_2}     & \imWV{DT_MF_2}      & \imWV{DT_FE-HPM_2}     & \imWV{DT_SR-D_2}        & \imWV{DT_MTF-GLP-HPM_2}        & \imWV{DT_MTF-GLP_2}       & \imWV{DT_AWLP_2} \\
MS &  \multicolumn{7}{c}{\footnotesize Reprojection error maps: defference between the MS and the reprojected pansharpening} \\[3mm]
\imWV{PAN_3} & \imWV{TV_3}        & \imWV{MF_3}         & \imWV{FE-HPM_3}        & \imWV{SR-D_3}           & \imWV{MTF-GLP-HPM_3}           & \imWV{MTF-GLP_3}         & \imWV{AWLP_3}    \\
PAN & TV & MF & FE-HPM & SR-D & ...-GLP-HPM & MTF-GLP & AWLP \\[1mm]
\imWV{MS_3}  & \imWV{DT_TV_3}     & \imWV{DT_MF_3}      & \imWV{DT_FE-HPM_3}     & \imWV{DT_SR-D_3}        & \imWV{DT_MTF-GLP-HPM_3}        & \imWV{DT_MTF-GLP_3}       & \imWV{DT_AWLP_3} \\
MS &  \multicolumn{7}{c}{\footnotesize Reprojection error maps: defference between the MS and the reprojected pansharpening} \\[3mm]
& R-$Q2^n$={\bf 0.986} & R-$Q2^n$=\underline{0.977} & R-$Q2^n$=0.976 & R-$Q2^n$=0.974 & R-$Q2^n$=0.974 & R-$Q2^n$=0.974 & R-$Q2^n$=0.973 \\
& $\DR$=0.227 & $\DR$=0.102 & $\DR$=0.095 & $\DR$=0.238 & $\DR$=\underline{0.079} & $\DR$={\bf 0.076} & $\DR$=0.097 \\
\end{tabular}
\caption{Sample crops from a WV2 sample image (Washington) and related full-resolution pansharpening. 
For each crop, in addition to the PAN and MS components (left-most column),
the best performing solutions according to R-$Q2^n$ are shown from left to right in decreasing order, next to the PAN.
For each method, both the R-$Q2^n$ and $\DR$ scores are reported on the bottom lines of the figure.
Moreover, next to the MS, for each pansharpening result the corresponding reprojection error map is shown.}
\label{fig:WV2FRcrops}
\end{figure}

\section{Conclusions}
\label{sec:conclusions}
To cope with the limitations of the reference-based reduced-resolution procedures for the quality assessment of pansharpening methods,
in this work we propose a new full-resolution no-reference evaluation framework.
By following Wald's protocol \cite{Wald97},
the full-resolution assessment must be carried out checking the ``consistency'' of the fused products with the MS and PAN input components rather than the ``synthesis'' capacity
which would require the availability of GTs.
In particular,
inspired by Khan's index \cite{Khan2009},
we have proposed the use of reprojection-based indexes with embedded alignment to handle mis-registered datasets for the assessment of the spectral consistency
between the pansharpened image and the input MS.
Besides,
the spatial consistency between the fused image and the PAN is quantified by averaging, spatially and spectrally, 
the fine-scale local correlation of individual super-resolved bands with the high-resolution PAN.

A key qualifying aspect of the proposed indexes is the absence of any resolution downgrading of the input data,
which frees the assessment from the effect of scale-dependent phenomena.
Experiments on reduced resolution datasets show that the reprojection indexes are reliable predictors of image quality as quantified by reference-based indexes
supporting their use in the full-resolution domain.
On the other hand, experiments on full-resolution data make clear 
that the local correlation-based index provides indications on image quality that largely agree with the judgement of human experts.
The proposed approach can be readily generalized to fusion tasks other than pansharpening
such as, for example, the combination of low-resolution hyperspectral and high-resolution multispectral images.

On the other hand,
the User must also be aware of some limitations. 
In particular, 
the reprojection requires the knowledge of an accurate estimation of the sensor MTF for a correct low-pass filtering.
A wrong estimation would lead to inaccurate spectral consistency assessment,
a problem shared with preexisting solutions.
Moreover,
the proposed correlation distortion index is based on the assumption that the PAN correlates well with all spectral bands.
Such a hypothesis is globally acceptable but there can be rare cases, spatially and spectrally well localized, for which it results too strong.

In order to ensure reproducible research,
the code for the proposed indexes can be found at
\url{https://github.com/matciotola/fr-pansh-eval-tool/}.

\bibliographystyle{plain}
\bibliography{refs}  






\end{document}